\documentclass[]{TEAI}
\usepackage{helvet}

\usepackage{amsmath}
\usepackage{natbib}
\usepackage{graphicx}
\usepackage{subcaption}

\usepackage[toc,page,header]{appendix}
\usepackage[utf8]{inputenc}
\usepackage[T1]{fontenc}
\usepackage{hyperref}
\usepackage{url}
\usepackage{booktabs}
\usepackage{amsfonts}
\usepackage{nicefrac}
\usepackage{xcolor}
\usepackage{wrapfig}
\usepackage{bm}
\usepackage{tikz}
\usepackage{array}
\usepackage{longtable}
\usepackage{pdflscape}
\usepackage{enumitem}
\usepackage{amssymb}
\usepackage{fontawesome}
\usepackage{titletoc}
\usepackage{comment}
\usepackage{tabularx}
\usepackage{minitoc}
\usepackage{etoolbox}

\definecolor{lightblue}{RGB}{200, 230, 255}
\definecolor{headerblue}{RGB}{150, 200, 255}
\definecolor{fwdbg}{RGB}{230,242,255}
\definecolor{fwdborder}{RGB}{120,165,220}
\definecolor{invbg}{RGB}{255,240,222}
\definecolor{invborder}{RGB}{220,160,90}

\usepackage{pgfplots}
\usepackage{float}
\usepackage{multirow}
\usepackage{makecell}
\usepackage{siunitx}
\usepackage[export]{adjustbox}
\usepackage{ragged2e}
\usepackage{caption}
\usepackage{pifont}
\usepackage[hang,flushmargin]{footmisc}
\usepackage{tcolorbox}
\tcbuselibrary{breakable}
\tcbuselibrary{skins}
\usepackage{listings}

\usepackage{pifont}
\usepackage{tikz}
\usepackage{enumitem}
\usepackage{wrapfig}
\usepackage{booktabs}
\usepackage{array}
\newcommand{\cmark}{\textcolor[RGB]{45,110,20}{\ding{51}}}
\newcommand{\xmark}{\textcolor[RGB]{200,0,0}{\ding{55}}}

\usepackage{subcaption}

\newcolumntype{C}[1]{>{\centering\arraybackslash}m{#1}}
\usepackage{makecell}

\title{OpenViTac: Learning and Benchmarking Visuo-Tactile Policies in a Unified Sim-and-Real Framework}

\author{
    Yifan Wu$^{1,2,\ast}$,
    Qin Li$^{1,2,3,\ast}$,
    Nan Min$^{1,2,\ast}$,
    Guojin Zhong$^{1,2,\dagger}$,
    Haoyu Zhao$^{4,\dagger}$,
    Zhiyuan Li$^{1,2}$,
    Houze Xu$^{1,2}$,
    Shengqi Xu$^{1,2}$,
    Xingyao Lin$^{1,2}$,
    Zijie Diao$^{1,2}$,
    Zhaoxiang Liu$^{6}$,
    Shiguo Lian$^{6}$,
    Shunlin Lu$^{5}$,
    Shihao Zhao$^{5}$,
    Ziyi Ye$^{1,2}$,
    Zuxuan Wu$^{1,2,5}$,
    Yu-Gang Jiang$^{1,2}$
}

\affiliation[1]{\mbox{Fudan University}}
\affiliation[2]{\mbox{Shanghai Key Laboratory of Multimodal Embodied AI}}
\affiliation[3]{\mbox{Hefei University of Technology}}
\affiliation[4]{\mbox{National University of Singapore}}
\affiliation[5]{\mbox{Neote AI}}
\affiliation[6]{\mbox{China Unicom}}

\abstract{
Tactile feedback provides embodied agents with physical information beyond visual observations, enabling more reliable interaction with the real world.
However, despite the rapid progress of vision-tactile-language-action (VTLA) policies, there remains a lack of unified benchmarks for evaluating tactile-enabled robot manipulation across simulation and the real world.
To address this gap, we introduce \textbf{OpenViTac}, a visuo-tactile manipulation benchmark for evaluating robot policies across simulation and the real world.
OpenViTac organizes contact-rich manipulation into four tactile-relevant capability dimensions and provides paired simulation--real-world settings for consistent evaluation of VLA, WAM, and VTLA policies.
Building upon this benchmark, we investigate how different tactile representations and integration strategies affect the performance of pretrained VLA models.
Correspondingly, we introduce \textbf{OpenVTLA}, a tactile augmentation framework that combines the best-performing representation and integration strategy.
Furthermore, we leverage the paired benchmark setting to study sim--real co-training and analyze factors affecting cross-domain policy learning.
Together, OpenViTac provides a unified platform for evaluating and advancing visuo-tactile robot manipulation.
}

\checkdata[Keywords]{Embodied AI, Visuo-Tactile, Simulation and Real}
\checkdata[Website]{\url{https://fvl-repo.github.io/OpenViTac/}}

\begin{document}
\maketitle

\begingroup
\renewcommand{\thefootnote}{\fnsymbol{footnote}}
\footnotetext[1]{Equal contribution.\quad
    $^{\dagger}$Project leaders..}
\endgroup

\begin{figure*}[t]
    \centering
    \includegraphics[width=\textwidth]{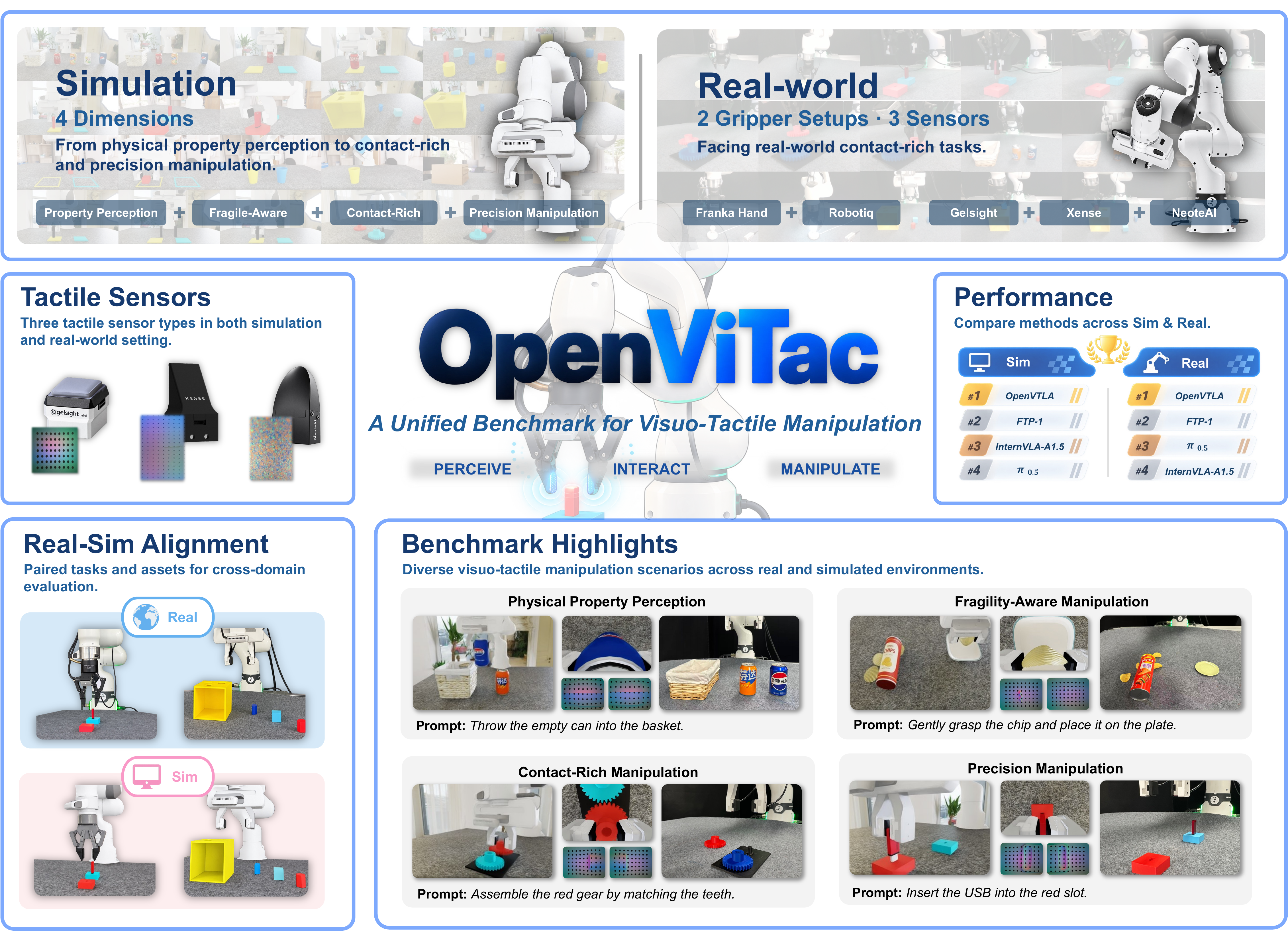}
    \caption{
    \textbf{Overview of OpenViTac.}
    OpenViTac establishes a unified benchmark for visuo-tactile manipulation, spanning four complementary capability dimensions: physical-property perception, fragility-aware interaction, contact-rich manipulation, and precision manipulation.
    It provides task-corresponding simulation and real-world settings with aligned assets, supports multiple gripper configurations and tactile sensors, and enables consistent evaluation of visuo-tactile policies across both domains.
    }
    \label{fig:teaser}
\end{figure*}

\vspace{-3mm}
\section{Introduction}

\begin{quote}
\centering
\vspace{1.5pt}
\hspace{-1cm}
\textit{``The ideas of sight and touch make two species entirely distinct and heterogeneous.''}\\[2pt]
\hspace{11cm}\textnormal{--- George Berkeley}
\vspace{1.5pt}
\end{quote}

Recent advances in Vision-Language-Action (VLA) models~\citep{brohan2023rt2,kim2024openvla} have enabled robots to leverage large-scale visual and linguistic knowledge for increasingly generalizable manipulation~\citep{intelligence2025pi05}.
These foundation policies have demonstrated strong performance across diverse manipulation scenarios~\citep{liu2023libero,khazatsky2024droid}, yet remain primarily grounded in visual observations.
However, many real-world interactions require understanding physical states that are not directly observable from vision alone, such as object properties, contact dynamics, and interaction forces~\citep{feng2026anytouch2}.
By providing direct feedback from physical interaction, tactile sensing~\citep{dong2017improved,team2026n0_foundation} augments visual perception with complementary physical information, enabling embodied agents to better understand and interact with the physical world, such as contact geometry, force cues, and material-dependent responses.
This has motivated a shift from purely vision-based policies toward Vision-Tactile-Language-Action (VTLA) models~\citep{huang2025tactilevla,yuan2026ftp,niu2026t-rex}, which combine global scene understanding from vision with local interaction feedback from touch.

Driven by the promise of tactile feedback, recent VTLA studies have rapidly expanded the design space of tactile-enabled robot policies, exploring different tactile representations~\citep{zhao2025t3,higuera2024sparsh,feng2026anytouch2}, fusion strategies~\citep{morissette2026tacfilm,xu2026ViTacMotor}, and policy architectures~\citep{yuan2026ftp,niu2026t-rex}.
Nevertheless, these approaches are typically evaluated under different sensors, tasks, robot embodiments, and experimental protocols, making systematic comparison challenging and leaving the contribution of tactile sensing insufficiently understood.
Existing manipulation benchmarks, such as LIBERO~\citep{liu2023libero}, RoboTwin~\citep{chen2025robotwin}, and RoboDojo~\citep{chen2026robodojo}, primarily focus on vision-based policies and do not support tactile sensing.
% While benchmarks such as LIBERO~\citep{liu2023libero}, RoboTwin~\citep{chen2025robotwin}, and RoboDojo~\citep{chen2026robodojo} have established scalable evaluation platforms with diverse manipulation tasks and simulation environments aligned with real-world robot settings, they are primarily designed around visual observations and do not explicitly evaluate the role of tactile feedback in manipulation.
Recent visuo-tactile benchmarks, such as Tabero~\citep{wu2026tabero} and UniVTAC~\citep{chen2026univtac}, have started to introduce tactile observations and contact-rich tasks for evaluation.
However, as shown in Tab.~\ref{tab:benchmark_comparison}, existing benchmarks still lack comprehensive coverage of tactile-relevant capabilities, diverse hardware configurations, and sim--real correspondence.

To address this, we introduce \textbf{OpenViTac}, a unified benchmark for evaluating visuo-tactile robot manipulation across simulation and the real world.
As shown in Fig.~\ref{fig:overview}, OpenViTac organizes tactile-enabled manipulation into four capability dimensions: physical property perception, fragility-aware manipulation, contact-rich manipulation, and precision manipulation, covering diverse scenarios that require understanding object properties, contact states, and fine-grained interaction dynamics.
We evaluate representative VLA, WAM, and VTLA policies~\citep{intelligence2025pi05,community2026starvla,zheng2026xvla,wu2026lingbot-vla-2,ma2026internvla-a1.5,team2026xiaomi,ali2025cosmos2.5,yuan2026fast-wam,li2026lingbot-va,team2026n_0-twam,yuan2026ftp,team2026n0-vtla} on OpenViTac, providing a comprehensive comparison of their tactile-enabled manipulation capabilities.

Building upon OpenViTac, we further investigate how tactile observations can be incorporated into vision-based policies through two design choices: tactile representation and tactile integration.
Our study reveals that stronger tactile representations and effective integration strategies consistently improve performance on contact-sensitive manipulation tasks.
Based on this analysis, we introduce \textbf{OpenVTLA}, a tactile augmentation framework for pretrained VLA models.
Quantitatively, the strongest VLA, WAM, and VTLA baselines achieve 51.5\%, 53.3\%, and 61.2\% average success rates in simulation, respectively, while OpenVTLA reaches 68.7\%.
The same trend holds in real-world evaluation, where OpenVTLA achieves 54.6\%, compared with 43.1\%, 48.5\%, and 51.9\% from the strongest VLA, WAM, and VTLA baselines.
Finally, leveraging the paired sim--real setting of OpenViTac, we investigate sim--real co-training as a strategy for improving real-world policy learning through additional simulation data.

Overall, our contributions are threefold:

\begin{table*}[t]
\centering
\caption{\textbf{Comparison of representative simulated manipulation benchmarks with OpenViTac.} We include recent visuo-tactile benchmarks together with representative vision-based benchmarks.}
\label{tab:benchmark_comparison}

\scriptsize
\setlength{\tabcolsep}{3.0pt}
\renewcommand{\arraystretch}{1.10}

% \resizebox{\textwidth}{!}{
\begin{tabular}{l|ccccccc}
\toprule
\textbf{Benchmark}
% & \makecell{\textbf{Gripper}\\\textbf{Types}}
& \textbf{Gripper Types}
& \makecell{\textbf{Tactile}\\\textbf{Sensing}}
& \makecell{\textbf{Tactile Sensor}\\\textbf{Types}}
& \makecell{\textbf{Property}\\\textbf{Perception}}
& \makecell{\textbf{Fragility-/Force-}\\\textbf{Aware}}
& \makecell{\textbf{Precision}\\\textbf{Manipulation}}
& \makecell{\textbf{Scene}\\\textbf{Diversity}} \\
\midrule
CALVIN~\citep{mees2022calvin} & 1 & \cmark & 1 & \xmark & \xmark & \xmark & \xmark \\
LIBERO~\citep{liu2023libero} & 1 & \xmark & - & \xmark & \xmark & \xmark & \xmark \\
ManiFeel~\citep{luu2025manifeel} & 1 & \cmark & 1 & \xmark & \xmark & \xmark & \xmark \\
Tabero~\citep{wu2026tabero} & 1 & \cmark & 1 & \xmark & \cmark & \xmark & \xmark \\
UniVTAC~\citep{chen2026univtac} & 1 & \cmark & 1 & \cmark & \xmark & \cmark & \xmark \\
NeoSim~\citep{team2026n0_foundation} & 1 & \cmark & 1 & \xmark & \cmark & \cmark & \xmark \\
SoftVTBench~\citep{jing2026softvtbench} & 1 & \cmark & 1 & \xmark & \cmark & \xmark & \xmark \\
\midrule
\textbf{OpenViTac (Ours)} & 2 & \cmark & 3 & \cmark & \cmark & \cmark & \cmark \\
\bottomrule
\end{tabular}
% }
\end{table*}

% Our contributions are threefold:
% \begin{itemize}
%     \item \textbf{A visuo-tactile benchmark.} We introduce OpenViTac, comprising 11 contact-rich manipulation tasks with task-specific success criteria, together with matched simulation and real-world settings for a subset of tasks that admit meaningful cross-domain correspondence.
    
%     \item \textbf{A systematic study of tactile integration.} We benchmark representative VLA, VTLA, and WAM policies and study how tactile signals should be incorporated into vision-based policies. We find that stronger tactile representations consistently improve contact-sensitive manipulation, leading to OpenVTLA, which achieves the best overall performance on our benchmark.
    
%     \item \textbf{A sim-real co-training study.} We jointly train vision-only and visuo-tactile policies on simulation and real-world data, showing consistent gains over single-domain training and highlighting action-state distribution alignment as a key factor for effective cross-domain learning.
% \end{itemize}

\begin{itemize}[leftmargin=1.5em, itemsep=1pt, topsep=1pt]
    \item \textbf{A unified visuo-tactile manipulation benchmark.}
    We introduce \textbf{OpenViTac}, a tactile-centric benchmark with four capability dimensions and paired simulation--real-world settings for evaluating VLA, WAM, and VTLA policies.

    \item \textbf {An investigation of tactile adaptation for pretrained VLAs.}
    We investigate tactile representation and integration strategies for adapting vision-based policies to touch, leading to \textbf{OpenVTLA}, a tactile augmentation framework achieving strong performance across simulation and real-world evaluations.

    \item \textbf{A sim--real co-training study.}
    We leverage OpenViTac's paired sim--real setting to investigate sim--real co-training and analyze the factors affecting cross-domain policy learning.
\end{itemize}

\section{Related Work}
\label{sec:related_work}

Here we review related work on visuo-tactile robot learning and visuo-tactile datasets and benchmarks. Additional discussion can be found in Appendix~\ref{app:additional_related_work}.

\subsection{Visuo-Tactile Robot Learning}

Recent robot foundation policies have largely developed around vision-language-action (VLA) models and world-action models (WAMs)~\citep{kim2024openvla,intelligence2025pi05}.
Tactile feedback provides complementary contact and physical-property information that can improve contact-sensitive manipulation, motivating tactile-augmented policies such as VTLA models~\citep{yuan2026ftp,team2026n0-vtla,niu2026t-rex,wang2026vitar} and tactile-aware world-action models~\citep{lou2026dreamtac,team2026n_0-twam}.
Existing approaches, however, differ substantially in how touch is incorporated, including feature modulation~\citep{morissette2026tacfilm}, action-level conditioning~\citep{team2026n0-vtla}, dedicated tactile experts~\citep{yuan2026ftp}, and late-stage action refinement~\citep{niu2026t-rex,wang2026vitar}.
Given the lack of consensus on how tactile information should be incorporated into pretrained VLAs, in this paper we systematically study representative tactile integration strategies under a shared $\pi_{0.5}$ backbone.

\subsection{Visuo-Tactile Datasets and Benchmarks}

General manipulation benchmarks such as LIBERO~\citep{liu2023libero}, RoboTwin~\citep{chen2025robotwin}, and RoboDojo~\citep{chen2026robodojo} provide standardized evaluation across diverse simulated tasks, but primarily focus on vision-based manipulation.
A broad range of tactile datasets has been developed for representation learning~\citep{yang2022touch,kerr2022self,higuera2024sparsh,feng2026anytouch2}, while tactile-enabled real-robot manipulation data have only recently begun to scale across tasks, sensors, and embodiments~\citep{yuan2026ftp,team2026n0_foundation,lou2026dreamtac}.
For policy-level evaluation, ManiFeel~\citep{luu2025manifeel} and TacO~\citep{zorin2026taco} study tactile sensing in downstream manipulation, while UniVTAC~\citep{chen2026univtac} and NeoSim~\citep{team2026n0_foundation} further provide systematic evaluation in simulation. 
However, existing simulation benchmarks still exhibit substantial discrepancies from real-world settings.
OpenViTac provides a unified tactile-centric benchmark with aligned sim--real tasks and additional real-world data across multiple tactile sensors.

\begin{figure}[!t]
    \centering

    % ---------- (a) Left ----------
    \begin{minipage}[c]{0.42\linewidth}
        \centering
        \includegraphics[width=0.94\linewidth]{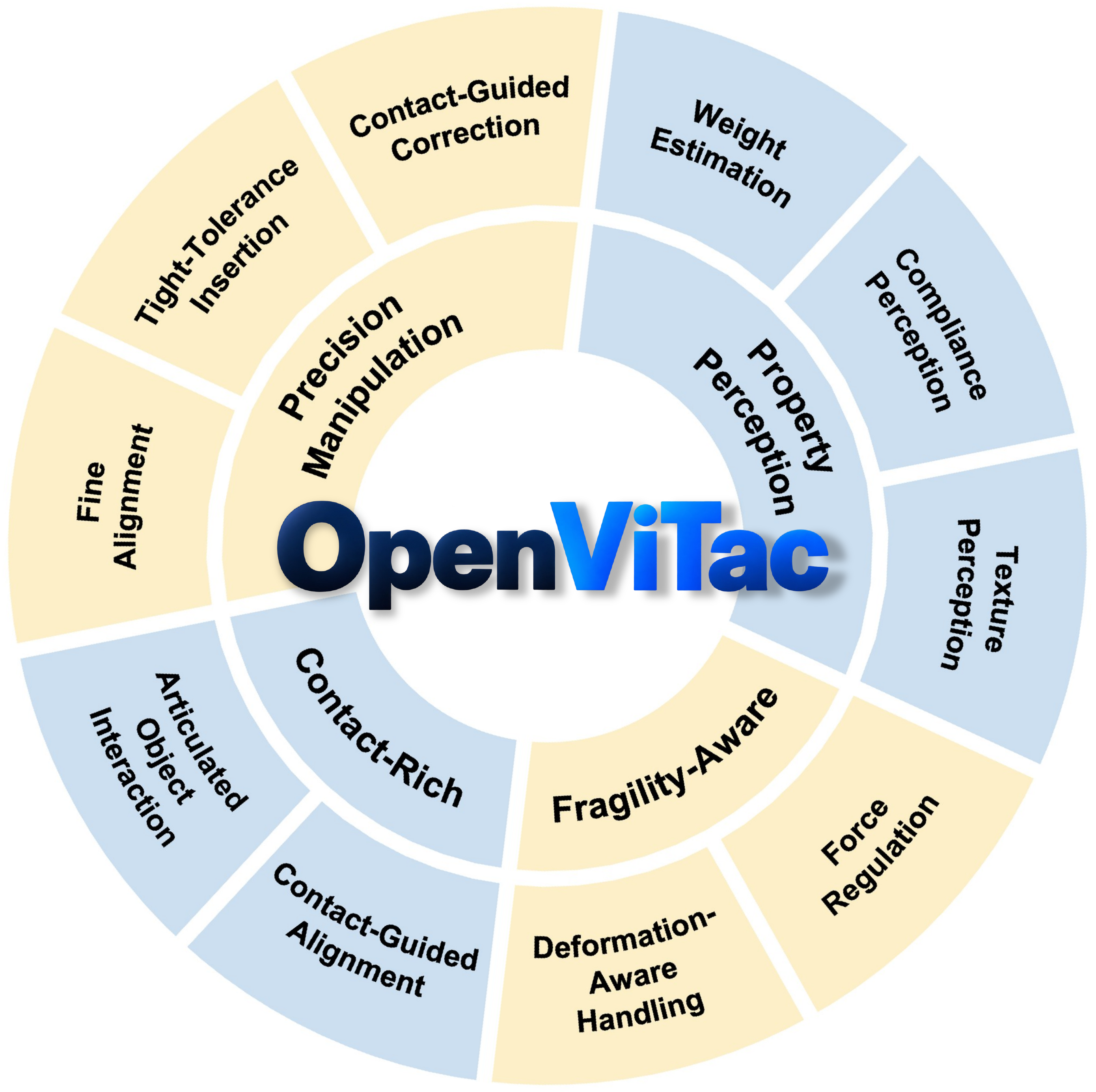}
        \vspace{0.6em}
        \subcaption{Capability Taxonomy}
        \label{fig:capability_taxonomy}
    \end{minipage}
    \hfill
    % ---------- (b)(c) Right ----------
    \begin{minipage}[c]{0.56\linewidth}
        \centering

        \includegraphics[width=0.9\linewidth]{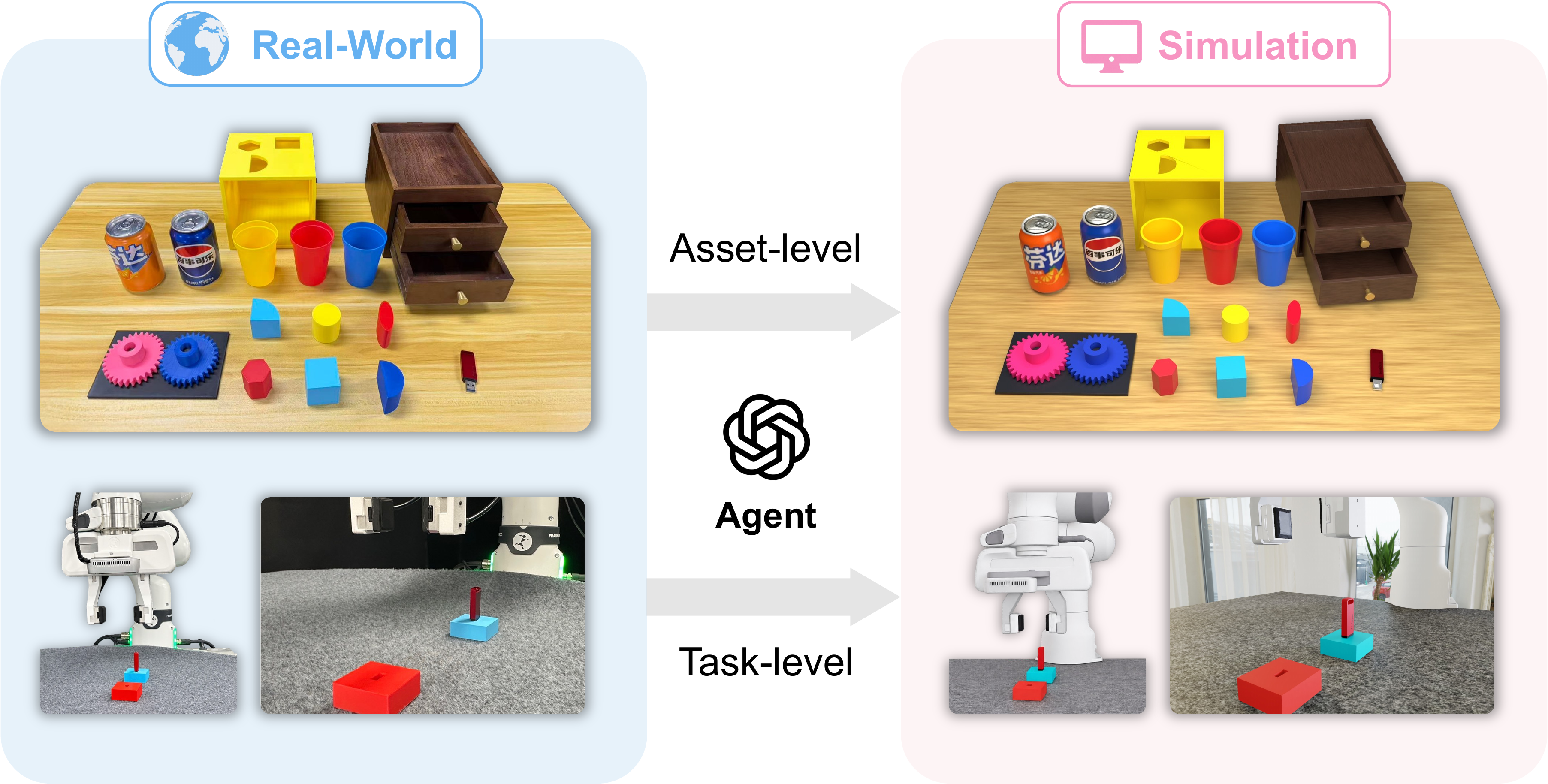}
        % \vspace{-0.1em}
        \subcaption{Real--Sim Alignment}
        \label{fig:real_sim_alignment}

        \vspace{0.5em}

        \includegraphics[width=\linewidth]{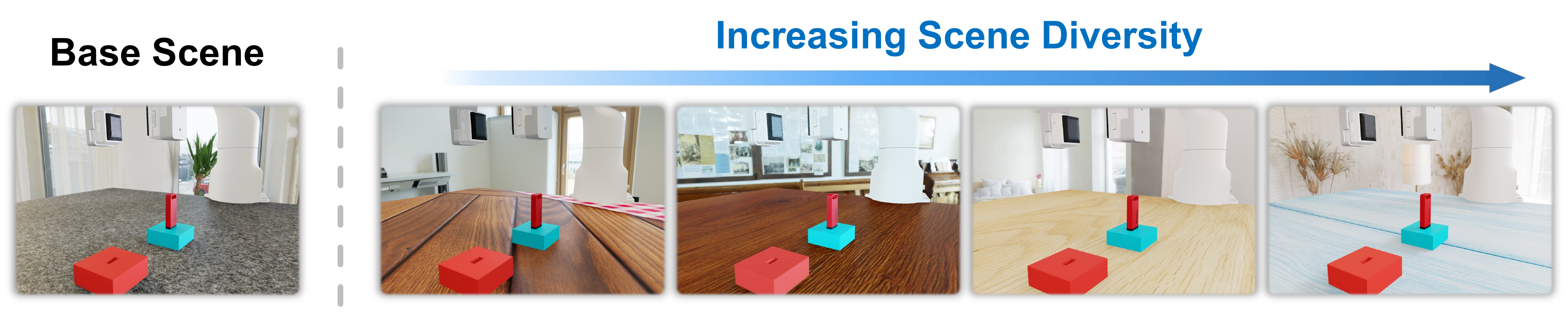}
        \vspace{-1.2em}
        \subcaption{Scene Diversity}
        \label{fig:scene_diversity}
    \end{minipage}

    \caption{
        \textbf{Overview of the OpenViTac design.}
        (a) Capability taxonomy with four visuo-tactile capabilities and their underlying skills.
        (b) Construction of task-corresponding simulation environments from real-world references at the asset- and task-levels.
        (c) Support for scene diversification through configurable appearance variations in simulation.
    }
    \label{fig:benchmark_design}
\end{figure}

\section{OpenViTac}
\label{sec:benchmark}

We introduce \textbf{OpenViTac}, a unified benchmark for evaluating visuo-tactile robot manipulation across simulation and the real world.
OpenViTac organizes tactile-enabled manipulation into four capability dimensions and provides 11 simulation tasks with 8 paired real-world tasks for systematic evaluation of VLA, WAM, and VTLA policies.
Building upon OpenViTac, we study tactile adaptation for pretrained VLA models through tactile representation and integration strategies, leading to \textbf{OpenVTLA}, and further investigate sim--real co-training enabled by the paired benchmark setting.

\subsection{Benchmark}
\label{sec:benchmark_design}

\textbf{Benchmark design.}
OpenViTac evaluates whether policies can exploit tactile feedback when visual observations provide incomplete information about physical interaction.
As illustrated in Fig.~\ref{fig:capability_taxonomy}, we organize the benchmark around four visuo-tactile capabilities: \textit{property perception}, \textit{fragility-aware}, \textit{contact-rich}, and \textit{precision}.
Further details are provided in Appendix~\ref{app:benchmark_details}.

\textbf{Task suite.}
We organize the tasks into four complementary visuo-tactile capabilities:
(1) \textit{Property Perception} (PP) evaluates whether tactile feedback can reveal visually ambiguous physical properties and support property-conditioned decisions.
It includes \textit{weight classification} (WC), \textit{hardness classification} (HC), and \textit{roughness classification} (RC) for property-conditioned placement; \textit{roughness-guided regrasp} (RR) for property-aware regrasping; and \textit{empty-can selection} (ECS) for distinguishing empty from filled cans through physical interaction.
(2) \textit{Fragility-Aware} (FA) evaluates force-sensitive manipulation of delicate objects.
\textit{Grasp chip} (GC) requires grasping and placing a fragile chip without damaging it.
(3) \textit{Contact-Rich} (CR) evaluates manipulation that requires sustained and effective physical contact.
It includes \textit{gear assembly} (GA), which requires maintaining gear engagement during assembly, and \textit{pull drawer} (PD), which requires maintaining a stable grasp while actuating an articulated drawer.
(4) \textit{Precision Manipulation} (PR) evaluates contact-guided manipulation under tight geometric tolerances.
It includes \textit{insert USB} (In-USB) and \textit{insert block v1} (In-B v1) for precise alignment and insertion, and \textit{insert block v2} (In-B v2), which additionally requires selecting the instructed object before insertion.
% \begin{itemize}[
%     labelindent=0pt,
%     leftmargin=*,
%     itemsep=0.1em,
%     topsep=0em
% ]
%     \item \textbf{Property Perception (PP).}
%     Evaluates whether tactile feedback can reveal visually ambiguous physical properties and support property-conditioned decisions.
%     It includes \textit{weight classification} (WC), \textit{hardness classification} (HC), and \textit{roughness classification} (RC) for property-conditioned placement; \textit{roughness-guided regrasp} (RR) for property-aware regrasping; and \textit{empty-can selection} (ECS) for distinguishing empty from filled cans through physical interaction.

%     \item \textbf{Fragility-Aware (FA).}
%     Evaluates force-sensitive manipulation of delicate objects.
%     \textit{Grasp chip} (GC) requires grasping and placing a fragile chip without damaging it.

%     \item \textbf{Contact-Rich (CR).}
%     Evaluates manipulation that requires sustained and effective physical contact.
%     It includes \textit{gear assembly} (GA), which requires maintaining gear engagement during assembly, and \textit{pull drawer} (PD), which requires maintaining a stable grasp while actuating an articulated drawer.

%     \item \textbf{Precision Manipulation(PR).}
%     Evaluates contact-guided manipulation under tight geometric tolerances.
%     It includes \textit{insert USB} (In-USB) and \textit{insert block v1} (In-B v1) for precise alignment and insertion, and \textit{insert block v2} (In-B v2), which additionally requires selecting the instructed object before insertion.
% \end{itemize}

\noindent\textbf{Real--sim alignment.}
As shown in Fig.~\ref{fig:real_sim_alignment}, we provide an agent-assisted real-to-sim pipeline for constructing task-corresponding simulation environments from real-world references.
(1) At the asset level, multi-view images of real-world objects are used to reconstruct their simulation counterparts, with the agent analyzing task-relevant structure and decomposing objects into functional parts when necessary;
(2) At the task level, reference object and target configurations together with robot initial states are used to instantiate the corresponding simulation task and construct executable sub-actions for trajectory collection.
OpenViTac further supports scene diversification by varying backgrounds, workspace appearance, and surrounding context while preserving the underlying task configuration, as illustrated in Fig.~\ref{fig:scene_diversity}, providing a flexible way to broaden the simulated data distribution for downstream training.

% \textbf{Real--sim alignment.}
% As shown in Fig.~\ref{fig:real_sim_alignment}, we provide an agent-assisted real-to-sim pipeline for constructing task-corresponding simulation environments from real-world references.
% \begin{itemize}[
%     labelindent=0pt,
%     leftmargin=*,
%     itemsep=0.1em,
%     topsep=0em
% ]
%     \item \textbf{Asset-level alignment.}
%     For each task, we capture multi-view images of the real-world objects and use an agent equipped with reconstruction skills to build their simulation counterparts.
%     The agent analyzes task-relevant object structure, decomposes objects into functional parts when necessary, and reconstructs the corresponding geometry and visual appearance.

%     \item \textbf{Task-level alignment.}
%     Real-world task instances provide reference object and target configurations together with robot initial states.
%     The agent uses these references to instantiate the corresponding simulation task and construct executable sub-actions.
%     Successful simulated trajectories are then collected through environment rollouts.
% \end{itemize}
% OpenViTac also supports scene diversification around the aligned base environments by varying backgrounds, workspace appearance, and surrounding context while preserving the underlying task configuration, as illustrated in Fig.~\ref{fig:scene_diversity}.
% This provides a flexible way to broaden the simulated data distribution for downstream training.

\textbf{Evaluation protocol.}
\label{sec:benchmark_evaluation}
We evaluate all policies under predefined and balanced task conditions using task-specific binary success criteria.
Further details are provided in Appendix~\ref{app:benchmark_details}.

\begin{figure}[!t]
    \centering
    \includegraphics[width=1.0\linewidth]{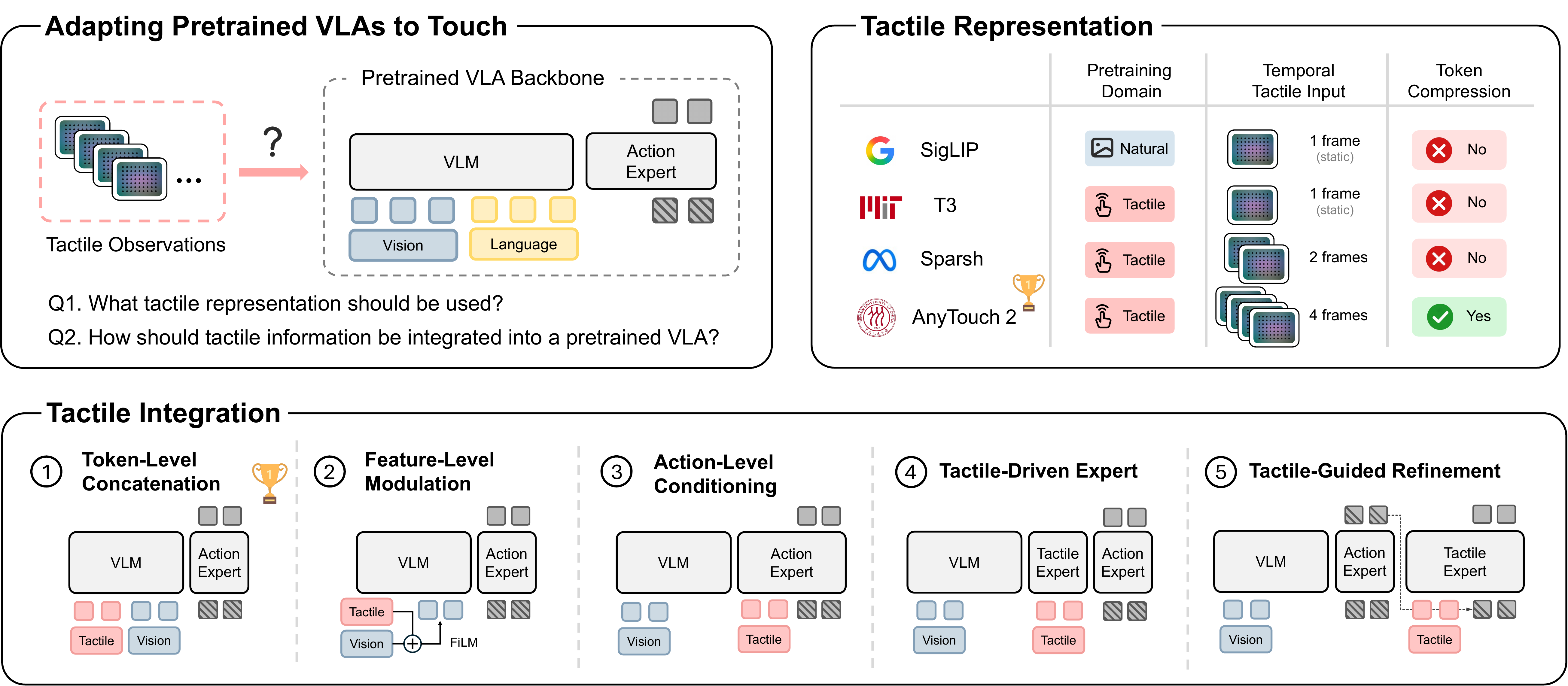}
    \caption{\textbf{Design space for adapting pretrained VLAs to touch.}
    We investigate two complementary dimensions: \textit{tactile representation} and \textit{tactile integration}.
    }
    \label{fig:tactile_design}
\end{figure}

\subsection{ADAPTING PRETRAINED VLAS TO TOUCH}
\label{sec:openvtla}

Large-scale visuo-tactile robot data remain limited and heterogeneous across sensors, embodiments, and manipulation tasks, making it difficult to train tactile-native policies at the scale of modern VLAs~\citep{yuan2026ftp,team2026n0-vtla}.
We therefore study a practical post-training setting in which all variants start from the same pretrained $\pi_{0.5}$~\citep{intelligence2025pi05} backbone, while varying how tactile observations are represented and incorporated into the policy.
As shown in Fig.~\ref{fig:tactile_design}, we investigate these choices along two dimensions: \textit{tactile representation} and \textit{tactile integration}.

% \textbf{Tactile Representation.}
% The most direct way to introduce tactile observations is to reuse SigLIP~\citep{zhai2023siglip}, the native visual encoder of $\pi_{0.5}$.
% We further consider tactile-specific representations: T3~\citep{zhao2025t3} operates on individual tactile frames, while Sparsh~\citep{higuera2024sparsh} and AnyTouch2~\citep{feng2026anytouch2} additionally incorporate temporal context to capture evolving contact information.
% Among them, AnyTouch2 will additionally compress the resulting representations into a compact token space.
% We compare these representations under the same token-level concatenation interface in Sec.~\ref{sec:tactile_analysis}; additional background is provided in Appendix~\ref{app:tactile_representation}.

\noindent\textbf{Tactile Representation.}
The first design choice is how tactile observations should be represented for a pretrained VLA.
The most direct option is to reuse SigLIP~\citep{zhai2023siglip}, the native visual encoder of $\pi_{0.5}$, which preserves the existing representation interface but is pretrained on natural images rather than tactile observations.
We therefore further consider recent tactile-specific representations, including T3~\citep{zhao2025t3}, Sparsh~\citep{higuera2024sparsh}, and AnyTouch2~\citep{feng2026anytouch2}.
T3 processes individual tactile frames, while Sparsh and AnyTouch2 incorporate short temporal histories to capture evolving contact information.
AnyTouch2 further adopts a compact tactile representation, consistent with the more localized and contact-specific information carried by tactile observations.
We compare these representations under the same token-level concatenation interface in Sec.~\ref{sec:tactile_analysis}; additional background is provided in Appendix~\ref{app:tactile_representation}.

% \textbf{Tactile Integration.}
% Given a tactile representation, we consider five representative integration strategies spanning different stages of the pretrained VLA, from early VLM fusion to late action refinement, as illustrated in Fig.~\ref{fig:tactile_design}.
% \begin{itemize}[
%     leftmargin=1.8em,
%     labelsep=0.4em,
%     topsep=0pt,
%     itemsep=0.2em,
%     parsep=0pt,
%     partopsep=0pt
% ]
%     \item[\ding{172}] \textbf{Token-Level Concatenation.}
%     Tactile tokens are concatenated with the vision--language sequence and processed jointly by the VLM.

%     \item[\ding{173}] \textbf{Feature-Level Modulation.}
%     Tactile features modulate intermediate visual representations through FiLM~\citep{morissette2026tacfilm} without introducing additional tactile tokens.

%     \item[\ding{174}] \textbf{Action-Level Conditioning.}
%     Tactile features bypass the VLM and directly condition the action expert, as in N0-VTLA~\citep{team2026n0-vtla}.

%     \item[\ding{175}] \textbf{Tactile-Driven Expert.}
%     A dedicated tactile expert processes tactile representations alongside the action expert, as in FTP-1~\citep{yuan2026ftp}.

%     \item[\ding{176}] \textbf{Tactile-Guided Refinement.}
%     Tactile information is introduced during later action-generation stages to refine partially generated actions, as in T-Rex~\citep{niu2026t-rex}.
% \end{itemize}

\noindent\textbf{Tactile Integration.}
Given a tactile representation, the second design choice is where and how tactile information enters the pretrained VLA.
As illustrated in Fig.~\ref{fig:tactile_design}, we consider five representative strategies:
\begin{itemize}[
    leftmargin=1.8em,
    labelsep=0.4em,
    topsep=0pt,
    itemsep=0.3em,
    parsep=0pt,
    partopsep=0pt
]
    \item[\ding{172}] \textbf{Token-Level Concatenation.}
    Tactile tokens are concatenated with the vision--language sequence, allowing tactile information to participate in VLM processing from the input stage.

    \item[\ding{173}] \textbf{Feature-Level Modulation.}
    Tactile features modulate intermediate visual representations through FiLM~\citep{perez2018film}, preserving the pretrained VLA pathway as much as possible~\citep{morissette2026tacfilm}.

    \item[\ding{174}] \textbf{Action-Level Conditioning.}
    Tactile features bypass the VLM and directly condition the action expert, leaving the vision--language backbone unchanged, as in N0-VTLA~\citep{team2026n0-vtla}.

    \item[\ding{175}] \textbf{Tactile-Driven Expert.}
    A dedicated tactile expert provides a separate computation pathway for tactile representations alongside the action expert, as in FTP-1~\citep{yuan2026ftp}.

    \item[\ding{176}] \textbf{Tactile-Guided Refinement.}
    Tactile information is introduced at later action-generation stages to refine partially generated actions rather than conditioning the full generation process, as in T-Rex~\citep{niu2026t-rex}.
\end{itemize}

% \noindent\textbf{Tactile Integration.}
% Given a tactile representation, the second design choice is where and how tactile information enters the pretrained VLA.
% As illustrated in Fig.~\ref{fig:tactile_design}, we consider five representative strategies.
% \ding{172}~\textit{Token-Level Concatenation} concatenates tactile tokens with the vision--language sequence, allowing tactile information to participate in VLM processing from the input stage.
% \ding{173}~\textit{Feature-Level Modulation} uses tactile features to modulate intermediate visual representations through FiLM~\citep{perez2018film}, keeping the policy structure closely aligned with the pretrained VLA while introducing tactile cues into the existing visual pathway~\citep{morissette2026tacfilm}.
% \ding{174}~\textit{Action-Level Conditioning} bypasses the VLM and directly conditions the action expert, leaving the vision--language backbone unchanged, as in N0-VTLA~\citep{team2026n0-vtla}.
% \ding{175}~\textit{Tactile-Driven Expert} introduces a dedicated tactile expert alongside the action expert to provide a separate computation pathway for tactile information, as in FTP-1~\citep{yuan2026ftp}.
% Finally, \ding{176}~\textit{Tactile-Guided Refinement} introduces tactile information at later stages of action generation to refine partially generated actions rather than conditioning the full generation process, as in T-Rex~\citep{niu2026t-rex}.

\textbf{OpenVTLA.}
Based on the above study, we instantiate OpenVTLA with AnyTouch2 and \ding{172} token-level concatenation.
AnyTouch2 encodes the tactile history into compact tokens, which are projected to the VLM hidden dimension and prepended to the visual token sequence.
This design introduces no additional fusion module or tactile-specific expert, preserving the original VLM and action-generation pathway of $\pi_{0.5}$.

%\subsection{Simulation-Real-World Co-training}
%\label{sec:method_cotrain}

%Moreover, we co-train OpenViTac-$\pi_{0.5}$ using simulated and real interaction data from the paired scenarios. Both sources contribute to policy training, while their scenario correspondence links the training data to common manipulation objectives. This connects joint learning to evaluation in the same set of simulated and physical environments.

%\textbf{Training setup.}
%The co-training study examines policy learning with data from both domains. Its experimental interpretation depends on the amount of simulated and real experience used and the training exposure allocated to each. In particular, the effect of adding simulated experience at a fixed real-data budget is distinct from the effect of changing the domain composition at a fixed total budget.

\subsection{Simulation--Real-World Co-training}
\label{sec:method_cotrain}

We further study sim--real co-training on OpenViTac, asking whether simulated demonstrations can improve the real-world performance of OpenVTLA under a fixed real-data budget.
For each task, we keep the real demonstrations $\mathcal{D}_{r}$ fixed and augment them with successful simulated demonstrations $\mathcal{D}_{s}$ collected across the diversified scene configurations in Fig.~\ref{fig:scene_diversity}.

\textbf{Sim-real data correspondence.} Following the task-aware digital-cousin principle~\citep{maddukuri2025sim}, we preserve the robot embodiment, task objectives, language instructions, object categories, and semantic success criteria across domains. We align action and proprioceptive representations, coordinate frames, units, gripper conventions, and control frequency, while approximately matching the real main and wrist camera viewpoints. Within this shared task structure, simulated scene diversity broadens the training distribution beyond the available real demonstrations.

\textbf{Mixed-domain training.}
We retain the OpenVTLA architecture and jointly train on real and simulated demonstrations with
\begin{equation}
    \mathcal{L}_{\mathrm{co}}
    =(1-\alpha)\mathcal{L}_{\mathrm{BC}}(\mathcal{D}_{r})
    +\alpha\mathcal{L}_{\mathrm{BC}}(\mathcal{D}_{s}),
\end{equation}
where $\mathcal{L}_{\mathrm{BC}}$ denotes the flow-matching behavior-cloning loss and $\alpha$ specifies the simulation sampling probability independently of dataset size. 
Training requires neither paired cross-domain trajectories nor an explicit feature-alignment loss. 
Real-only and co-trained policies share the same real demonstrations, architecture, pretrained initialization, and optimization budget. We also apply this protocol to vision-only $\pi_{0.5}$ to examine how tactile input affects the benefit of simulated data.

\section{Experiments}
\label{sec: experiments}

% We first benchmark representative VLA, WAM, and VTLA policies on OpenViTac in both simulation and the real world.
% We then investigate how tactile representation and integration affect the adaptation of pretrained VLAs, followed by a study of simulation--real-world co-training.
% Implementation and training details are provided in Appendix~\ref{app:policy_details}, with additional experimental results in Appendix~\ref{app:additional_experiments}.

\subsection{Experimental Setup}
\label{sec:exp_setup}

We benchmark representative policies from three model families on OpenViTac, including six VLA models, three WAM models, and one existing VTLA model, together with our OpenVTLA.
The VLA baselines include $\pi_{0.5}$~\citep{intelligence2025pi05}, StarVLA (Qwen3)~\citep{community2026starvla}, InternVLA-A1.5~\citep{ma2026internvla-a1.5}, XVLA~\citep{zheng2026xvla}, Lingbot-VLA-2~\citep{wu2026lingbot-vla-2}, and Xiaomi-Robotics~\citep{team2026xiaomi}.
For WAMs, we evaluate FastWAM~\citep{yuan2026fast-wam}, Lingbot-VA~\citep{li2026lingbot-va}, and StarVLA (Cosmos)~\citep{ali2025cosmos2.5}.
For VTLAs, we compare against FTP-1~\citep{yuan2026ftp}.

All policies are trained and evaluated in a single-task setting, with a separate policy instantiated for each benchmark task.
Evaluation follows predefined task conditions in both simulation and the real world; for tasks with multiple variants, each variant is evaluated separately, and the results are aggregated at the task level.
The reported Avg.\ is the unweighted mean of task-level success rates.
Implementation and training details for all evaluated policies are provided in Appendix~\ref{app:policy_details}.

\begin{table*}[!t]
\centering
\caption{Comparison of VLA, WAM, and VTLA models on OpenViTac \textbf{in the simulation environment}. 
% Avg. denotes the mean success rate across all tasks.
\textbf{Bold} denotes the best results and \underline{underline} denotes the second-best.}
\label{tab:sim_comparison}

\scriptsize
\setlength{\tabcolsep}{3.0pt}
\renewcommand{\arraystretch}{1.10}

% \resizebox{\textwidth}{!}{
\begin{tabular}{ll|cccccccccccc}
\toprule

\multicolumn{2}{c|}{\multirow{2}{*}{\textbf{Models}}}
& \multicolumn{5}{c}{\textbf{PP}}
& \multicolumn{1}{c}{\textbf{FA}}
& \multicolumn{2}{c}{\textbf{CR}}
& \multicolumn{3}{c}{\textbf{PR}}
& \multirow{2}{*}{\makebox[2.8em][c]{\textbf{Avg.}}} \\

\cmidrule(lr){3-7}
\cmidrule(lr){8-8}
\cmidrule(lr){9-10}
\cmidrule(lr){11-13}

\multicolumn{2}{c|}{}
& \makebox[2.8em][c]{\textbf{WC}}
& \makebox[2.8em][c]{\textbf{HC}}
& \makebox[2.8em][c]{\textbf{RC}}
& \makebox[2.8em][c]{\textbf{RR}}
& \makebox[2.8em][c]{\textbf{ECS}}
& \makebox[2.8em][c]{\textbf{GC}}
& \makebox[2.8em][c]{\textbf{GA}}
& \makebox[2.8em][c]{\textbf{PD}}
& \makebox[2.8em][c]{\textbf{In-USB}}
& \makebox[2.8em][c]{\textbf{In-B v1}}
& \makebox[2.8em][c]{\textbf{In-B v2}}
& \\

\midrule

\multirow{6}{*}{VLAs}
& $\pi_{0.5}$ & 53 & 60 & 64 & 52 & 48 & 39 & \underline{87} & \textbf{99} & 14 & 9 & \textbf{9.3} & 48.6 \\
& StarVLA (Qwen3) & 44 & 38 & 42 & 35 & 32 & 10 & 26 & 35 & 4 & 0 & 0 & 24.2 \\
& InternVLA-A1.5 & 53 & 76 & 65 & 77 & 47 & 53 & 79 & 98 & 3 & 6 & \textbf{9.3} & 51.5 \\
& XVLA & 28 & 30 & 56 & 47 & 45 & 42 & 84 & 76 & 2 & 3 & 0.7 & 37.6 \\
& Lingbot-VLA-2 & 52 & 58 & 58 & 53 & \underline{52} & 41 & 78 & 92 & 8 & 5 & 2.7 & 45.4 \\
& Xiaomi-Robotics & 52 & 46 & 54 & 62 & 38 & 24 & 73 & 84 & 5 & 2 & 0 & 40.0 \\

\midrule

\multirow{3}{*}{WAMs}
& FastWAM & 49 & 84 & 67 & 7 & 9 & 50 & 51 & 88 & 6 & 0 & 0 & 37.4 \\
& Lingbot-VA & 50 & 44 & 39 & 40 & 44 & 22 & 59 & 15 & 2 & 2 & 6.0 & 29.4 \\
& StarVLA (Cosmos) & 51 & 67 & 94 & 9 & 13 & 1 & 33 & 22 & 3 & 1 & 3.3 & 27.0 \\

\midrule

\multirow{2}{*}{VTLAs}
& FTP-1 & \underline{61} & \textbf{100} & \textbf{100} & \underline{78} & 50 & \underline{60} & 83 & 94 & \textbf{27} & \textbf{12} & 8.3 & \underline{61.2} \\
& OpenVTLA (Ours) & \textbf{89} & \underline{99} & \textbf{100} & \textbf{90} & \textbf{73} & \textbf{69} & \textbf{91} & \textbf{99} & \textbf{27} & \underline{11} & 8.0 & \textbf{68.7} \\

\bottomrule
\end{tabular}
% }
\end{table*}

\begin{table*}[!t]
\centering
\caption{Comparison of VLA, WAM, and VTLA models on OpenViTac \textbf{in the real-world environment}. 
% Avg. denotes the mean success rate across all tasks.
\textbf{Bold} denotes the best results and \underline{underline} denotes the second-best.}
\label{tab:real_comparison}
\scriptsize
\setlength{\tabcolsep}{4.0pt}
\renewcommand{\arraystretch}{1.10}

% \resizebox{\textwidth}{!}{%
\begin{tabular}{ll| *{9}{C{3.5em}}}
\toprule

\multicolumn{2}{c|}{\multirow{2}{*}{\textbf{Models}}}
& \multicolumn{2}{c}{\textbf{PP}}
& \multicolumn{1}{c}{\textbf{FA}}
& \multicolumn{2}{c}{\textbf{CR}}
& \multicolumn{3}{c}{\textbf{PR}}
& \multirow{2}{*}{\textbf{Avg.}} \\

\cmidrule(lr){3-4}
\cmidrule(lr){5-5}
\cmidrule(lr){6-7}
\cmidrule(lr){8-10}

\multicolumn{2}{c|}{}
& \textbf{RC}
& \textbf{ECS}
& \textbf{GC}
& \textbf{GA}
& \textbf{PD}
& \textbf{In-USB}
& \textbf{In-B v1}
& \textbf{In-B v2}
& \\

\midrule

\multirow{4}{*}{VLAs}
& $\pi_{0.5}$      & 10/20 & 10/20 & 16/20 & 7/20 & 12/20 & \underline{3/20} & \textbf{5/20} & 4/30 & 41.0 \\
& InternVLA-A1.5  & 10/20 & 10/20 & 17/20 & \underline{9/20} & \underline{13/20} & \underline{3/20} & 3/20 & \textbf{6/30} & 43.1 \\
& Lingbot-VLA-2   & 10/20 & 10/20 & 15/20 & 3/20 & 10/20 & 1/20 & 2/20 & \underline{5/30} & 34.0 \\
& Xiaomi-Robotics & 10/20 & 10/20 & 16/20 & 5/20 & 8/20 & 1/20 & 3/20 & 2/30 & 34.0 \\

\midrule

\multirow{1}{*}{WAMs}
& FastWAM         & 8/20 & 9/20 & 12/20 & 5/20 & 10/20 & 0/20 & 1/20 & 2/30 & 29.0 \\

\midrule

\multirow{2}{*}{VTLAs}
& FTP-1           & \underline{18/20} & \underline{14/20} & \textbf{19/20} & \underline{9/20} & \textbf{14/20} & \underline{3/20} & \underline{4/20} & 3/30 & \underline{51.9} \\
& OpenVTLA (Ours) & \textbf{19/20} & \textbf{17/20} & \textbf{19/20} & \textbf{10/20} & 12/20 & \textbf{4/20} & 3/20 & \underline{5/30} & \textbf{54.6} \\

\bottomrule
\end{tabular}%
% }
\end{table*}

\subsection{Main Results}
\label{sec:exps_results}

\textbf{Evaluation in simulation.}
The simulation results in Tab.~\ref{tab:sim_comparison} reveal clear performance differences across policy families.
The evaluated WAMs generally underperform the stronger VLA baselines, while VTLA policies achieve the strongest overall performance.
The benefits of tactile sensing are particularly evident on \textit{Property Perception} (PP) and \textit{Fragility-Aware} (FA) tasks, where successful execution requires understanding physical properties and regulating contact interactions.
On \textit{Contact-Rich} (CR) tasks, strong VLAs remain competitive, indicating that visual planning and reliable action execution remain important even with tactile feedback.
In contrast, \textit{Precision Manipulation} (PR) remains challenging across all model families, highlighting the difficulty of fine-grained contact control.
Overall, OpenVTLA achieves the highest benchmark average of $68.7\%$, outperforming FTP-1 and the strongest VLA baseline by $7.5$ and $17.2$ percentage points, respectively.

\textbf{Evaluation in the real world.}
The real-world results in Tab.~\ref{tab:real_comparison} exhibit a consistent trend with simulation.
VLA baselines achieve average success rates ranging from $34.0\%$ to $43.1\%$, while FastWAM obtains $29.0\%$.
In comparison, FTP-1 and OpenVTLA achieve $51.9\%$ and $54.6\%$, respectively, demonstrating a clear overall advantage for tactile-enabled policies.
The largest improvements appear on PP tasks, where tactile feedback provides direct information about latent physical properties, while FA and CR tasks show more moderate gains.
PR remains challenging and exhibits mixed results across policy families, suggesting that tactile sensing alone is insufficient for reliable fine-grained manipulation.
Overall, OpenVTLA achieves the highest real-world average success rate of $54.6\%$.

\textbf{Sim--real consistency.}
Following SIMPLER~\citep{li2024simpler}, we assess whether simulation preserves relative policy performance rather than requiring identical success rates across domains.
Across seven methods with complete results on the eight shared tasks, the task-averaged scores yield Pearson $r=0.967$ and MMRV $=0.006$, indicating strong agreement between simulation and real-world evaluation.
OpenVTLA and FTP-1 rank first and second in both domains.
These results support OpenViTac as a useful simulation proxy for comparative policy evaluation.

\textbf{Tactile benefits depend on how touch is incorporated.}
On the eight tasks shared across simulation and the real world, OpenVTLA improves over $\pi_{0.5}$ by $13.6$ and $13.5$ percentage points, respectively.
However, the performance gap between FTP-1 and OpenVTLA indicates that access to tactile observations alone does not determine policy performance.
This motivates the controlled study in Sec.~\ref{sec:tactile_analysis}, where we examine how tactile representation and integration affect the adaptation of pretrained VLAs to touch.

\subsection{Understanding Tactile Adaptation in Pretrained VLAs}
\label{sec:tactile_analysis}

\begin{wrapfigure}{r}{0.3\linewidth}
    \centering
    \vspace{-0.8em}
    \includegraphics[width=0.95\linewidth]{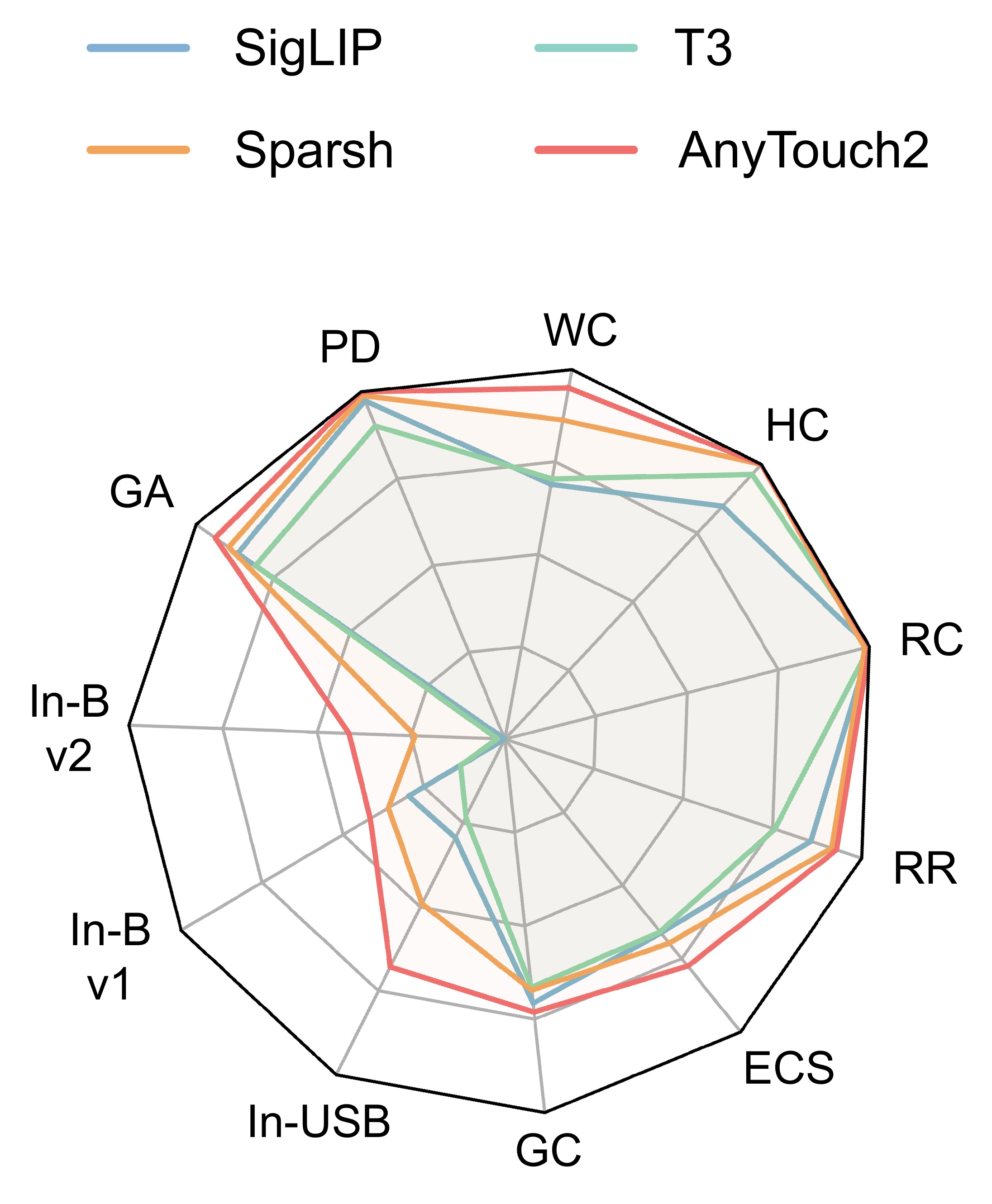}
    \vspace{-1em}
    \caption{
    Performance of different tactile representations under the same token-level concatenation strategy.}
    \label{fig:encoder_radar}
    \vspace{-0.8em}
\end{wrapfigure}
\textbf{What tactile representation is effective?}
We first compare different tactile representations under the same token-level concatenation interface.
As shown in Fig.~\ref{fig:encoder_radar}, the choice of tactile representation has a substantial impact on downstream performance.
After normalizing task-level results to percentages, AnyTouch2 achieves the highest average score of 68.7\%, outperforming Sparsh (64.7\%), SigLIP (59.5\%), and T3 (58.1\%).
The advantage is particularly pronounced on tasks that require reasoning about evolving contact states; for example, AnyTouch2 improves over Sparsh by 12 percentage points on ECS and 10 points on USB insertion.
More generally, the temporally-aware representations, Sparsh and AnyTouch2, tend to outperform the frame-wise alternatives, suggesting that tactile dynamics provide useful information beyond static contact appearance.
We further examine two factors underlying this trend:
(a) replacing the pretrained AnyTouch2 encoder with a randomly initialized counterpart leads to a clear performance degradation, indicating that tactile-specific pretraining provides useful representations for VLA adaptation;
(b) removing temporal variation by replacing the tactile history with repeated copies of the current frame particularly degrades physical-property recognition, including hardness and weight discrimination.
These controlled studies support the importance of both tactile pretraining and temporal dynamics; detailed results are provided in Appendix~\ref{app:additional_experiments}.

\begin{wraptable}{r}{0.55\linewidth}
    \centering
    \vspace{-0.8em}
    \caption{
    \textbf{Tactile integration comparison.}
    Scores are averaged over four task groups.
    ${\pi_{0.5}}^{\dagger}$ denotes variants obtained by adapting the pretrained
    $\pi_{0.5}$ backbone with different tactile integration strategies.
    \textbf{Bold} denotes the best results and \underline{underline} denotes the second-best.
    }
    \label{tab:tactile_integration}

    \scriptsize
    \setlength{\tabcolsep}{2.8pt}
    \renewcommand{\arraystretch}{1.10}

    \begin{tabular}{lcc|cccc|c}
    \toprule
    \textbf{Method}
    & \textbf{Representation}
    & \textbf{Integration}
    & \makebox[2.1em][c]{\textbf{PP}}
    & \makebox[2.1em][c]{\textbf{FA}}
    & \makebox[2.1em][c]{\textbf{CR}}
    & \makebox[2.1em][c]{\textbf{PR}}
    & \makebox[2.1em][c]{\textbf{Avg.}} \\
    \midrule

    OpenVTLA
    & AnyTouch2
    & \ding{172} Concat
    & \textbf{90.2}
    & \textbf{69.0}
    & \textbf{95.0}
    & \underline{15.3}
    & \textbf{68.7} \\

    ${\pi_{0.5}}^{\dagger}$
    & Sparsh
    & \ding{173} Modulation
    & \underline{82.2}
    & 49.0
    & 73.0
    & 5.0
    & 56.5 \\

    ${\pi_{0.5}}^{\dagger}$
    & AnyTouch2
    & \ding{174} Action
    & 70.8
    & 43.0
    & 62.0
    & 4.7
    & 48.6 \\

    FTP-1
    & T3
    & \ding{175} Expert
    & 77.8
    & \underline{60.0}
    & \underline{88.5}
    & \textbf{15.8}
    & \underline{61.2} \\

    ${\pi_{0.5}}^{\dagger}$
    & AnyTouch2
    & \ding{176} Refine
    & 63.0
    & 51.0
    & 81.0
    & 8.7
    & 50.4 \\

    \bottomrule
    \end{tabular}

    \vspace{-0.4em}
\end{wraptable}
\textbf{Where should tactile information be integrated?}
We next compare representative tactile integration strategies in Fig.~\ref{fig:tactile_design}, with results summarized across PP, FA, CR, and PR in Tab.~\ref{tab:tactile_integration}.
Overall, \ding{172}~token-level concatenation achieves the highest average performance and leads on PP, FA, and CR, while remaining competitive on PR.
Among the AnyTouch2-based variants, introducing tactile information directly at the VLM input substantially outperforms \ding{174}~action-level conditioning and \ding{176}~late refinement, suggesting that early access to touch is beneficial for learning contact-dependent representations and decisions.
A possible explanation is that token-level concatenation allows tactile information to interact with the full pretrained VLM throughout policy computation, rather than restricting it to downstream action-generation stages.
Together with the representation study above, these results support a simple recipe for tactile adaptation of pretrained VLAs: use a pretrained temporally-aware tactile encoder and inject its tokens directly into the VLM input sequence.

\subsection{Simulation--Real-World Co-training}
\label{sec:exp_cotrain}

We study whether simulated demonstrations can improve real-world performance under a fixed real-data budget on Insert USB (In-USB) and Gear Assembly (GA), using both vision-only $\pi_{0.5}$ and OpenVTLA.
For each task, all policies are trained with $100$ real demonstrations, with co-training additionally incorporating either $250$ or $500$ simulated demonstrations collected across diversified scene configurations.
We further evaluate the real-only and $500$-simulation policies under lighting perturbations.
Each model--task--condition setting is evaluated over $20$ trials.

\begin{wraptable}{r}{0.45\linewidth}
    \centering
    \vspace{-1.5em}
    \caption{
    \textbf{Sim--real co-training and robustness.}
    All settings use 100 real demonstrations per task; ``Sim.'' denotes the number of additional simulated demonstrations.
    }
    \label{tab:cotrain}

    \scriptsize
    \setlength{\tabcolsep}{3pt}
    \renewcommand{\arraystretch}{1.10}

    % \resizebox{\linewidth}{!}{
    \begin{tabular}{lc|cc|cc}
        \toprule
        \multirow{2}{*}{\textbf{Model}}
        & \multirow{2}{*}{\makebox[2.6em][c]{\textbf{Sim.}}}
        & \multicolumn{2}{c|}{\textbf{Std. Lighting}}
        & \multicolumn{2}{c}{\textbf{Lighting Perturb.}} \\
        \cmidrule(lr){3-4}
        \cmidrule(lr){5-6}
        &
        & \makebox[3.2em][c]{\textbf{In-USB}}
        & \makebox[3.2em][c]{\textbf{GA}}
        & \makebox[3.2em][c]{\textbf{In-USB}}
        & \makebox[3.2em][c]{\textbf{GA}} \\
        \midrule

        \multirow{3}{*}{$\pi_{0.5}$}
        & 0   & 3/20 & 7/20  & 0/20 & 2/20 \\
        & 250 & 5/20 & 9/20  & --   & --   \\
        & 500 & 8/20 & 10/20 & 4/20 & 8/20 \\
        \midrule

        \multirow{3}{*}{OpenVTLA}
        & 0   & 4/20 & 10/20 & 2/20 & 10/20 \\
        & 250 & 6/20 & 10/20 & --   & --    \\
        & 500 & 8/20 & 12/20 & 6/20 & 12/20 \\

        \bottomrule
    \end{tabular}
    % }

    \vspace{-0.8em}
\end{wraptable}
\textbf{Diverse simulated data improves real-world performance.}
As shown in Tab.~\ref{tab:cotrain}, adding simulated demonstrations improves both policies without increasing the real-data budget. With $500$ simulated demonstrations, $\pi_{0.5}$ improves from $15\%$ to 
$40\%$ on In-USB and from $35\%$ to $50\%$ on GA. OpenVTLA improves from $20\%$ to $40\%$ and from $50\%$ to $60\%$, respectively. Averaged over the two tasks, these correspond to gains of $20$ percentage points for $\pi_{0.5}$ and $15$ points for OpenVTLA. The smaller gain for OpenVTLA may partly reflect residual differences in contact-induced tactile signals across domains, motivating consistent sensor calibration and improved tactile simulation.

\textbf{Co-training and touch improve performance under lighting perturbations.}
Under perturbed lighting, real-only $\pi_{0.5}$ falls to $0\%$ on In-USB and $10\%$ on GA. Adding $500$ simulated demonstrations raises these results to $20\%$ and $40\%$, showing that broader simulated training supports better performance under visual disturbance, although degradation relative to standard lighting remains. OpenVTLA is less sensitive to the perturbation: its real-only policy achieves $10\%$ on In-USB and retains $50\%$ on GA, while co-training yields $30\%$ and $60\%$. In particular, its GA success rate is unchanged by lighting perturbations under both training settings. This pattern is consistent with tactile feedback providing contact information that is less sensitive to ambient lighting, complementing the robustness gained from diverse simulated experience.

% \section{Conclusion}

% In this work, we introduce \textbf{OpenViTac}, a visuo-tactile manipulation benchmark that enables systematic evaluation of tactile-enabled robot policies across simulation and the real world.
% By organizing manipulation scenarios into four tactile-relevant capability dimensions, OpenViTac provides a unified testbed for comparing VLA, WAM, and VTLA policies under consistent settings.
% Building upon this benchmark, we further investigate how tactile observations can be incorporated into pretrained VLA models through different representation and integration strategies, leading to \textbf{OpenVTLA}, a tactile augmentation framework that achieves strong performance across both simulation and real-world evaluations.
% We further explore sim--real co-training enabled by the paired benchmark setting, demonstrating the potential of leveraging simulated experience to improve real-world manipulation.
% Despite these advances, several challenges remain for future visuo-tactile manipulation benchmarks.
% In particular, the correspondence between simulated and real tactile observations is still limited by the fidelity of tactile rendering, sensor modeling, and multimodal calibration.
% Future work could further improve simulation realism through more accurate sensor and camera calibration, while expanding benchmark coverage toward more complex tactile-centric interactions, such as deformable object manipulation, surface interaction, and long-horizon household tasks.

\section{Conclusion, Limitations and Future Work}

\textbf{Conclusion.}
We introduced \textbf{OpenViTac}, a visuo-tactile manipulation benchmark for systematically evaluating tactile-enabled robot policies across simulation and the real world.
OpenViTac organizes manipulation into four tactile-relevant capability dimensions and provides a unified testbed for comparing VLA, WAM, and VTLA policies.
Building on this benchmark, we studied tactile representation and integration strategies for pretrained VLA models and introduced \textbf{OpenVTLA}, which achieves strong performance in both simulation and real-world evaluations.
We further explored sim--real co-training, showing that additional simulated experience can improve real-world manipulation performance.

\textbf{Limitations and future work.}
OpenViTac still faces a sim--real gap in tactile observations due to limitations in tactile rendering, sensor modeling, and multimodal calibration.
Future work will improve sensor and camera calibration and extend the benchmark toward richer tactile-centric interactions, including deformable-object manipulation, surface interaction, and long-horizon household tasks.
We also plan to support more diverse robot embodiments, including dexterous hands, to broaden the evaluation of visuo-tactile manipulation under more complex contact dynamics.

\clearpage

\bibliographystyle{plainnat}
\bibliography{main}

\clearpage

\clearpage
\appendix
\setcounter{table}{0}
\setcounter{figure}{0}
\renewcommand{\thetable}{S\arabic{table}}
\renewcommand{\thefigure}{S\arabic{figure}}

\begin{center}
{\Large\bf Appendix}
\end{center}

\section*{Overview}

This appendix provides additional details, experimental results, and visualizations that complement the main paper.
The appendix is organized as follows:
\begin{itemize}[leftmargin=1.5em, itemsep=2pt, topsep=2pt]
    \item \textbf{Additional Related Work (Appendix~\ref{app:additional_related_work}):}
    reviews related research on tactile learning and manipulation.
    
    \item \textbf{Benchmark and Task Details (Appendix~\ref{app:benchmark_details}):}
    describes the tasks, sensing platforms, and success criteria.

    \item \textbf{Policy Implementation and Training Details (Appendix~\ref{app:policy_details}):}
    documents policy implementations, tactile inputs, and training settings.

    \item \textbf{Additional Experimental Results (Appendix~\ref{app:additional_experiments}):}
    reports supplementary results and focused ablations.
    
    \item \textbf{Task Visualizations (Appendix~\ref{app:task_visualizations}):}
    illustrates tasks and tactile observations.

\end{itemize}

\section{Additional Related Work}
\label{app:additional_related_work}

\subsection{Tactile Representation Learning}
\label{app:tactile_representation}

Early tactile representation learning often exploits paired visual--tactile observations to learn cross-modal features.
Touch and Go~\citep{yang2022touch} supports self-supervised learning from human-collected visual--tactile interactions, while SSVTP~\citep{kerr2022self} and Visuo-Tactile Transformers~\citep{chen2022visuo} further explore cross-modal representations for perception and manipulation.
Subsequent work shifts toward more transferable tactile representations across sensors and tasks.
UniTouch~\citep{yang2024binding} aligns heterogeneous tactile observations with pretrained visual embeddings, Sparsh~\citep{higuera2024sparsh} learns general-purpose tactile features through self-supervised pretraining, UniT~\citep{xu2025unit} adopts generative reconstruction, and T3~\citep{zhao2025t3} targets transfer across heterogeneous tactile sensors and downstream tasks.
More recent methods increasingly exploit the temporal structure of touch: AnyTouch2~\citep{feng2026anytouch2} models dynamic and force-aware tactile information over short observation histories, while motion-based approaches further leverage tactile deformation dynamics to characterize evolving contact states~\citep{xu2026ViTacMotor}.
These developments motivate our comparison of generic visual, tactile-specific, and temporally-aware representations under a common policy interface.

\subsection{Simulation and Real-World Robot Learning}
\label{app:sim_real_related}

Simulation provides a scalable way to generate interaction data and evaluate robot policies.
Recent tactile platforms and benchmarks, including Tabero~\citep{wu2026tabero}, SoftVTBench~\citep{jing2026softvtbench}, UniVTAC~\citep{chen2026univtac}, and NeoSim~\citep{team2026n0_foundation}, further extend simulation to tactile manipulation.
However, their simulated tasks are generally not constructed as direct counterparts of real-world settings, leaving differences in objects, scene layouts, task configurations, and initial conditions.
A complementary line of work reduces such discrepancies through real-to-sim alignment: SIMPLER~\citep{li2024simpler} calibrates simulated evaluation against real-world policy behavior, RialTo~\citep{torne2024rialto} reconstructs digital twins for policy improvement, and SimFoundry~\citep{ranawaka2026simfoundry} automates the construction of task-relevant digital cousins.
These studies highlight the importance of preserving task-relevant correspondence between simulation and the real world rather than reproducing physical scenes exactly.
Recent studies further show that appropriately aligned simulated data can complement real demonstrations through sim--real co-training~\citep{maddukuri2025sim,lei2026mechanistic}.
Motivated by these findings, we construct aligned simulated and physical tactile-manipulation tasks and study whether simulated demonstrations can improve real-world visuo-tactile policy learning.

\section{Benchmark and Task Details}
\label{app:benchmark_details}
\label{app:setup}

\subsection{Benchmark Scope and Task Taxonomy}
\label{app:task_taxonomy}

OpenViTac comprises 11 visuo-tactile manipulation tasks for evaluating robot policies.
Following the taxonomy introduced in Sec.~\ref{sec:benchmark}, we organize these tasks into four capability categories: physical property perception, fragility-aware manipulation, contact-rich manipulation, and precision manipulation.
Tab.~\ref{tab:task_taxonomy} summarizes the task abbreviations and the primary capability evaluated by each task.

\begin{table}[H]
    \centering
    \small
    \setlength{\tabcolsep}{4pt}
    \caption{
    OpenViTac task taxonomy.
    Simulation and real-world evaluations use the same task semantics and
    success criteria for tasks instantiated in both domains.
    }
    \label{tab:task_taxonomy}
    \begin{tabularx}{\textwidth}{@{}p{0.17\textwidth} p{0.075\textwidth} p{0.20\textwidth} X@{}}
        \toprule
        Category & Abbr. & Task & Capability under evaluation \\
        \midrule

        \makecell[l]{Physical Property\\Perception}
        & WC & Weight Classify
        & Infer object weight through interaction and place the object at the corresponding target. \\

        & HC & Hardness Classify
        & Infer object hardness through contact and place the object at the corresponding target. \\

        & RC & Roughness Classify
        & Infer surface roughness through contact and place the object at the corresponding target. \\

        & RR & Roughness-Guided Regrasp
        & Infer relative roughness and keep or regrasp the appropriate object accordingly. \\

        & ECS & Empty-Can Selection
        & Distinguish visually similar empty and filled cans from their physical properties and select the instructed one. \\

        \midrule

        \makecell[l]{Fragility-Aware\\Manipulation}
        & GC & Grasp Chip
        & Grasp and place a fragile chip while avoiding damage throughout the manipulation. \\

        \midrule

        \makecell[l]{Contact-Rich\\Manipulation}
        & GA & Gear Assembly
        & Mount and rotate a gear while maintaining engagement with the paired gear. \\

        & PD & Pull Drawer
        & Maintain a stable grasp on the handle while pulling the articulated drawer open. \\

        \midrule

        \makecell[l]{Precision\\Manipulation}
        & In-USB & Insert USB
        & Align and insert a USB connector into the target opening under tight geometric tolerance. \\

        & In-B v1 & Insert Block v1
        & Align and insert a block into the corresponding target opening. \\

        & In-B v2 & Insert Block v2
        & Select the instructed object and then align and insert it into the corresponding target opening. \\

        \bottomrule
    \end{tabularx}
\end{table}

\subsection{Simulation and Real-world Environment}
\label{app:simulation_setup}

OpenViTac builds scalable visuo-tactile simulation environments with corresponding real-world task instantiations for consistent policy evaluation.
The simulation environments are developed upon NVIDIA Isaac Sim 5.1.0 and Isaac Lab 2.3.0, with tactile observations generated using TacEx-based tactile simulation~\citep{nguyen2024tacex}.
The platform supports three representative tactile sensors, including GelSight, Xense, and NeoteAI, together with two widely used gripper configurations: the Franka Hand and Robotiq gripper.

To generate realistic visuo-tactile observations, OpenViTac separates visual asset construction from tactile simulation.
While TacEx provides physics-based tactile rendering through contact interaction, it does not directly model the full visual appearance of objects, such as diverse textures and colors.
Therefore, we construct visually realistic object assets independently and associate them with TacEx-compatible tactile representations using shared object geometry.
This design enables consistent RGB and tactile observations from the same simulated objects.

For real-world evaluation, we instantiate the corresponding tasks using the same robot embodiments, gripper configurations, tactile sensors, and task objectives as in simulation.
Such correspondence preserves consistent interaction semantics between simulation and reality, enabling meaningful comparison of policies across both environments.

To facilitate policy deployment and comparison across different robot learning frameworks, OpenViTac provides standardized robot state and action interfaces.
Both end-effector-space and joint-space representations are supported, including multiple rotation parameterizations such as quaternions, rotation matrices, and rotation vectors.
These representations are consistently provided in both simulation and real-world environments, enabling compatibility with pretrained manipulation policies that adopt different action and state formulations.

\subsection{Data Collection Statistics}
\label{app:data_statistics}

OpenViTac contains both simulated trajectories and real-world robot demonstrations collected under three tactile sensing configurations: GelSight, Xense, and Neote.
The collected data statistics are summarized in Table~\ref{tab:data_statistics}.

For each tactile sensor configuration, the simulation benchmark contains 1,300 episodes across 11 tasks, corresponding to approximately 918K recorded frames in total.
The real-world dataset contains 2,400 demonstrations per tactile sensor configuration across 8 tasks, with approximately 2.1M recorded frames collected in total.

To improve task diversity, several tasks include multiple object configurations or language-conditioned variations.
For example, Insert Block v2 contains three object semantics with two scene layouts, resulting in six data collection configurations in the real-world setting.
Similarly, simulation tasks with binary physical properties, such as weight, hardness, and roughness classification, are balanced across different object conditions.

\begin{table}[H]
    \centering
    \footnotesize
    \setlength{\tabcolsep}{10pt}
    \caption{Data collection statistics of OpenViTac.
    Statistics are reported per tactile sensor configuration for both simulated episodes and real-world demonstrations.}
    \label{tab:data_statistics}

    \begin{subtable}[c]{0.48\textwidth}
        \centering
        \caption{Simulation}
        \begin{tabular}{@{}lr@{}}
            \toprule
            Task & Episodes \\
            \midrule
            Weight Classify (WC) & 100 \\
            Hardness Classify (HC) & 100 \\
            Roughness Classify (RC) & 100 \\
            Roughness-Guided Regrasp (RR) & 100 \\
            Empty-Can Selection (ECS) & 100 \\
            \addlinespace[2pt]
            Grasp Chip (GC) & 100 \\
            Gear Assembly (GA) & 100 \\
            Pull Drawer (PD) & 100 \\
            \addlinespace[2pt]
            Insert USB (In-USB) & 100 \\
            Insert Block v1 (In-B v1) & 100 \\
            Insert Block v2 (In-B v2) & 300 \\
            \midrule
            \textbf{Total} & \textbf{1,300} \\
            \bottomrule
        \end{tabular}
    \end{subtable}
    \hfill
    \begin{subtable}[c]{0.48\textwidth}
        \centering
        \caption{Real-world}
        \begin{tabular}{@{}lr@{}}
            \toprule
            Task & Demonstrations \\
            \midrule
            Roughness Classify (RC) & 100 \\
            Empty-Can Selection (ECS) & 100 \\
            Grasp Chip (GC) & 100 \\
            Gear Assembly (GA) & 200 \\
            Pull Drawer (PD) & 100 \\
            Insert USB (In-USB) & 200 \\
            Insert Block v1 (In-B v1) & 400 \\
            Insert Block v2 (In-B v2) & 1200 \\
            \midrule
            \textbf{Total} & \textbf{2,400} \\
            \bottomrule
        \end{tabular}
    \end{subtable}

\end{table}

\subsection{Full Task Catalog}
\label{app:task_catalog}

Table~\ref{tab:task_catalog} provides detailed descriptions of the 11 manipulation tasks in OpenViTac.
Each task is designed to evaluate a specific tactile-relevant capability while maintaining consistent task objectives across simulation and real-world settings.
The success conditions are defined based on task completion requirements, including object states, interaction outcomes, and physical constraints when applicable.

\begin{table}[htbp]
    \centering
    \footnotesize
    \setlength{\tabcolsep}{4pt}
    \renewcommand{\arraystretch}{1.08}
    \caption{
    Task catalog of OpenViTac, including task objectives and high-level success conditions.
    }
    \label{tab:task_catalog}

    \begin{tabularx}{\textwidth}{@{}
        >{\raggedright\arraybackslash}p{0.21\textwidth}
        >{\raggedright\arraybackslash}p{0.35\textwidth}
        >{\raggedright\arraybackslash}X
    @{}}
        \toprule
        \textbf{Task} &
        \textbf{Task Objective} &
        \textbf{Success Condition} \\
        \midrule

        Weight Classify (WC) &
        Infer whether a block is light or heavy through interaction
        and place it on the corresponding target. &
        The block is correctly classified and placed on the corresponding target plate. \\
        \addlinespace[3pt]

        Hardness Classify (HC) &
        Infer whether a block is soft or hard through contact
        interaction and place it on the corresponding target. &
        The block is correctly classified and placed on the corresponding target plate. \\
        \addlinespace[3pt]

        Roughness Classify (RC) &
        Infer whether a block is smooth or rough through contact
        interaction and place it on the corresponding target. &
        The block is correctly classified and placed on the corresponding target plate. \\
        \addlinespace[3pt]

        Empty-Can Selection (ECS) &
        Identify the empty can among visually similar cans through
        physical interaction and place it into the basket. &
        The empty can is correctly identified and successfully placed into the basket. \\
        \addlinespace[3pt]

        Roughness-Guided Regrasp (RR) &
        Determine whether to retain the initially grasped block
        based on roughness and regrasp when necessary. &
        The rough block is correctly selected and placed on the target plate. \\
        \addlinespace[3pt]

        Grasp Chip (GC) &
        Grasp and transport a fragile chip while avoiding damage
        during manipulation. &
        The chip is successfully grasped, transported, and placed
        without damage while satisfying physical safety constraints. \\
        \addlinespace[3pt]

        Gear Assembly (GA) &
        Assemble a gear onto the base and rotate it to drive the
        paired gear through physical engagement. &
        The gears are successfully assembled and achieve coupled
        rotation through stable mechanical interaction. \\
        \addlinespace[3pt]

        Pull Drawer (PD) &
        Grasp the drawer handle and pull the drawer open while
        maintaining stable contact. &
        The drawer is successfully grasped and pulled open with
        the required displacement. \\
        \addlinespace[3pt]

        Insert USB (In-USB) &
        Align the USB connector with the target slot and complete
        insertion. &
        The USB connector is successfully aligned and inserted into
        the target slot. \\
        \addlinespace[3pt]

        Insert Block v1 (In-B v1) &
        Grasp the designated cube and insert it into the matching
        opening under different scene layouts. &
        The designated block is successfully aligned and inserted
        into the corresponding opening. \\
        \addlinespace[3pt]

        Insert Block v2 (In-B v2) &
        Select the language-specified object and insert it into the
        corresponding opening across different layouts. &
        The instructed object is correctly selected and successfully
        inserted into the matching opening. \\

        \bottomrule
    \end{tabularx}
\end{table}

\newpage
\section{Policy Implementation and Training Details}
\label{app:policy_details}

We train a separate policy for each task.
For simulation, we train on 100 demonstration episodes per task, except Insert Block v2, which uses 300 episodes.
For real-world training, we randomly select 100 episodes per task from the larger collected demonstration pool, or 300 episodes for Insert Block v2.
The evaluated VLA, WAM, and VTLA policies are listed in Tab.~\ref{tab:baseline_training_configuration}.
For previously published models, we use each method's prescribed initialization and retain its official training configuration except for the training steps and global batch size specified in the table.
FastWAM, Lingbot-VA, and StarVLA (Cosmos) follow the LIBERO configuration in their respective official repositories.
OpenVTLA uses the same implementation as $\pi_{0.5}$.
Model-specific preprocessing and action representations follow the corresponding source recipe, with adapters used to train and execute each policy on the benchmark tasks.

\begin{table}[htbp]
    \centering
    \footnotesize
    \caption{Training settings for the evaluated policies.}
    \label{tab:baseline_training_configuration}
    \setlength{\tabcolsep}{7pt}
    \begin{tabular}{@{}cl|cc@{}}
        \toprule
        \multicolumn{2}{c|}{Models} & Steps & Global batch \\
        \midrule
        \multirow{6}{*}{VLAs}
        & $\pi_{0.5}$ & 20k & 64 \\
        & StarVLA (Qwen3) & 20k & 64 \\
        & InternVLA-A1.5 & 40k & 32 \\
        & XVLA & 30k & 64 \\
        & Lingbot-VLA-2 & 20k & 64 \\
        & Xiaomi-Robotics & 20k & 64 \\
        \midrule
        \multirow{3}{*}{WAMs}
        & FastWAM & 20k & 8 \\
        & Lingbot-VA & 5k & 8 \\
        & StarVLA (Cosmos) & 20k & 8 \\
        \midrule
        \multirow{2}{*}{VTLAs}
        & FTP-1 & 20k & 64 \\
        & OpenVTLA (Ours) & 20k & 64 \\
        \bottomrule
    \end{tabular}
\end{table}

Training steps count optimizer updates, and global batch size includes any gradient accumulation.
The native optimizer, learning-rate schedule, and prediction horizon follow the source recipe unless adaptation to the policy's action representation is required.
Because the architectures and training recipes differ across families, cross-family results compare complete policies rather than isolating the effect of a single training choice.
We evaluate the final checkpoint of each training run instead of selecting a checkpoint by validation performance.

\section{Additional Experimental Results}
\label{app:additional_experiments}

We further examine two factors that may contribute to the effectiveness of tactile representations: tactile-specific pretraining and temporal context. All variants retain the same OpenVTLA architecture and token-level integration interface, changing only the factor under study.

\textbf{Effect of tactile pretraining.} 
Tab.~\ref{tab:tactile_pretraining} compares the pretrained AnyTouch2 encoder with an identically structured encoder initialized from scratch.
Removing tactile pretraining reduces the overall performance from $68.7\%$ to $56.5\%$, with consistent degradation across all four capability groups.
The largest drop occurs on Fragility-Aware (FA) tasks, while substantial decreases are also observed on Property Perception (PP) and Contact-Rich (CR) tasks.
These results indicate that tactile-specific pretraining provides representations that transfer effectively to downstream VLA adaptation.

\begin{table}[h]
    \centering
    \caption{
    Comparison between pretrained AnyTouch2 and its randomly initialized counterpart across the four capability groups.}
    \label{tab:tactile_pretraining}

    \footnotesize
    \setlength{\tabcolsep}{5pt}
    \renewcommand{\arraystretch}{1.10}

    \begin{tabular}{l|cccc|c}
        \toprule
        \textbf{Representation} & \textbf{PP} & \textbf{FA} & \textbf{CR} & \textbf{PR} & \textbf{Avg.} \\
        \midrule
        AnyTouch2 (Random Init.) & 78.5 & 34.0 & 82.0 & 10.2 & 56.5 \\
        AnyTouch2 (Pretrained)   & \textbf{90.2} & \textbf{69.0} & \textbf{95.0} & \textbf{15.3} & \textbf{68.7} \\
        \bottomrule
    \end{tabular}
\end{table}

\textbf{Effect of temporal context.}
We next remove temporal variation while preserving the encoder and policy architecture by replacing the tactile history with repeated copies of the current frame.
As shown in Tab.~\ref{tab:tactile_temporal}, this decreases the overall performance from $68.7\%$ to $58.4\%$.
The degradation is most pronounced on PP and FA tasks, where observations of evolving contact can provide information about physical properties and interaction states.
Similar trends are also observed when introducing temporal tactile context to SigLIP and T3, suggesting that the benefit of temporal information is not specific to AnyTouch2.
Overall, these results indicate that tactile dynamics provide useful information beyond static contact appearance, particularly for tasks that depend on physical-property and contact-state estimation.

\begin{table}[h]
    \centering
    \caption{
    Comparison between the original tactile history and repeated copies of the current frame across the four capability groups.}
    \label{tab:tactile_temporal}

    \footnotesize
    \setlength{\tabcolsep}{5pt}
    \renewcommand{\arraystretch}{1.10}

    \begin{tabular}{l|cccc|c}
        \toprule
        \textbf{Temporal Input} & \textbf{PP} & \textbf{FA} & \textbf{CR} & \textbf{PR} & \textbf{Avg.} \\
        \midrule
        Repeated Current Frame & 79.0 & 52.0 & 90.0 & 5.2 & 58.4 \\
        Tactile History        & \textbf{90.2} & \textbf{69.0} & \textbf{95.0} & \textbf{15.3} & \textbf{68.7 }\\
        \bottomrule
    \end{tabular}
\end{table}

\clearpage
\section{Task Visualizations}
\label{app:task_visualizations}

\subsection{Visualizations Across Tactile Sensors}

\begin{figure}[H]
  \centering
  \includegraphics[width=0.8\linewidth]{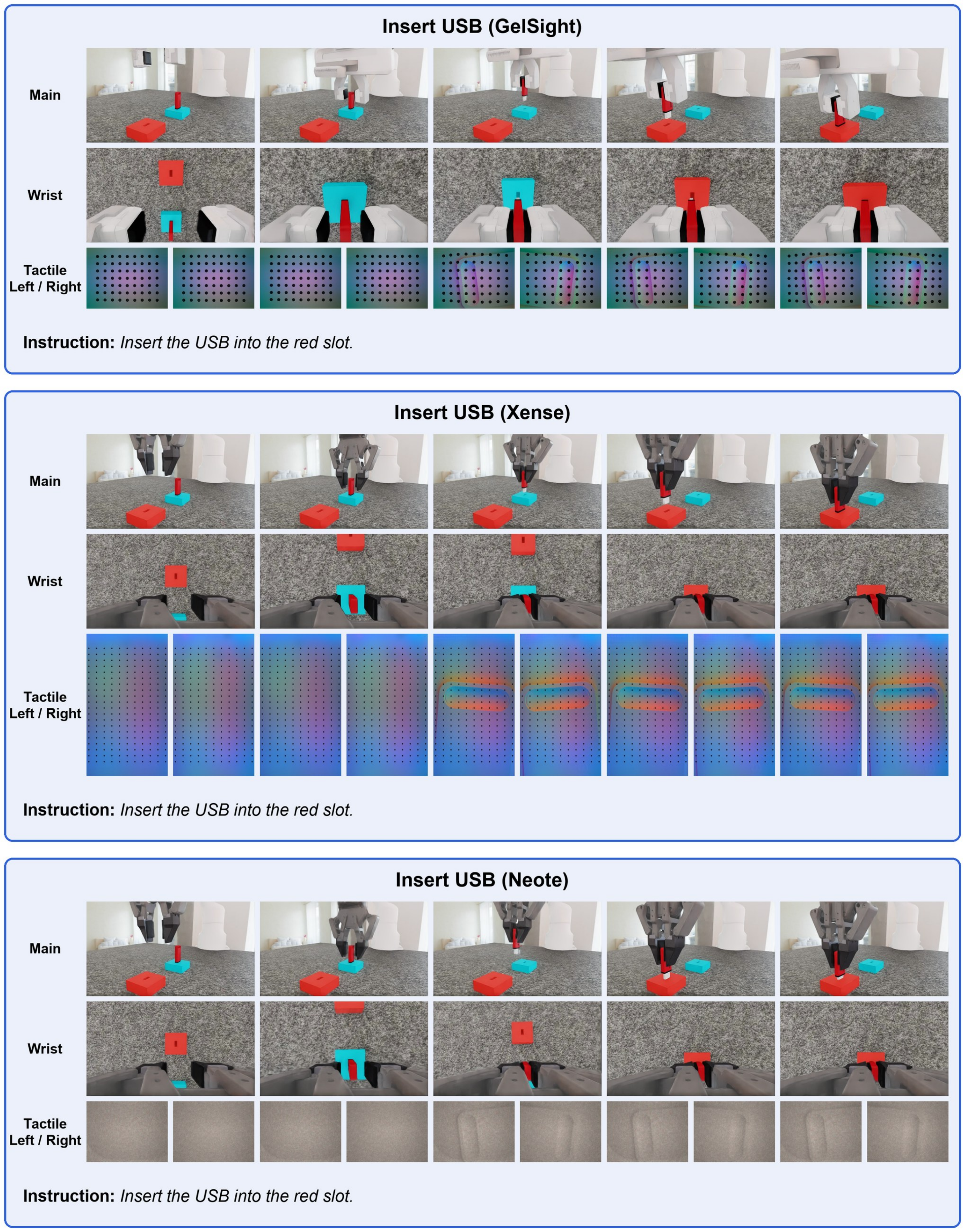}
  \caption{Visualization of the \textit{insert USB} task using GelSight, Xense, and NeoteAI tactile sensors.}
  \label{fig:three_sensors_visualization}
\end{figure}

Fig.~\ref{fig:three_sensors_visualization} compares representative observations from the insert USB task with GelSight, Xense, and NeoteAI sensors.
Each panel shows camera views together with left- and right-fingertip tactile readings at selected stages of the interaction.
The comparison illustrates how the same task objective appears through different tactile sensing modalities.

\clearpage
\subsection{Simulated Tasks}

\begin{figure}[H]
  \centering
  \includegraphics[width=0.8\linewidth]{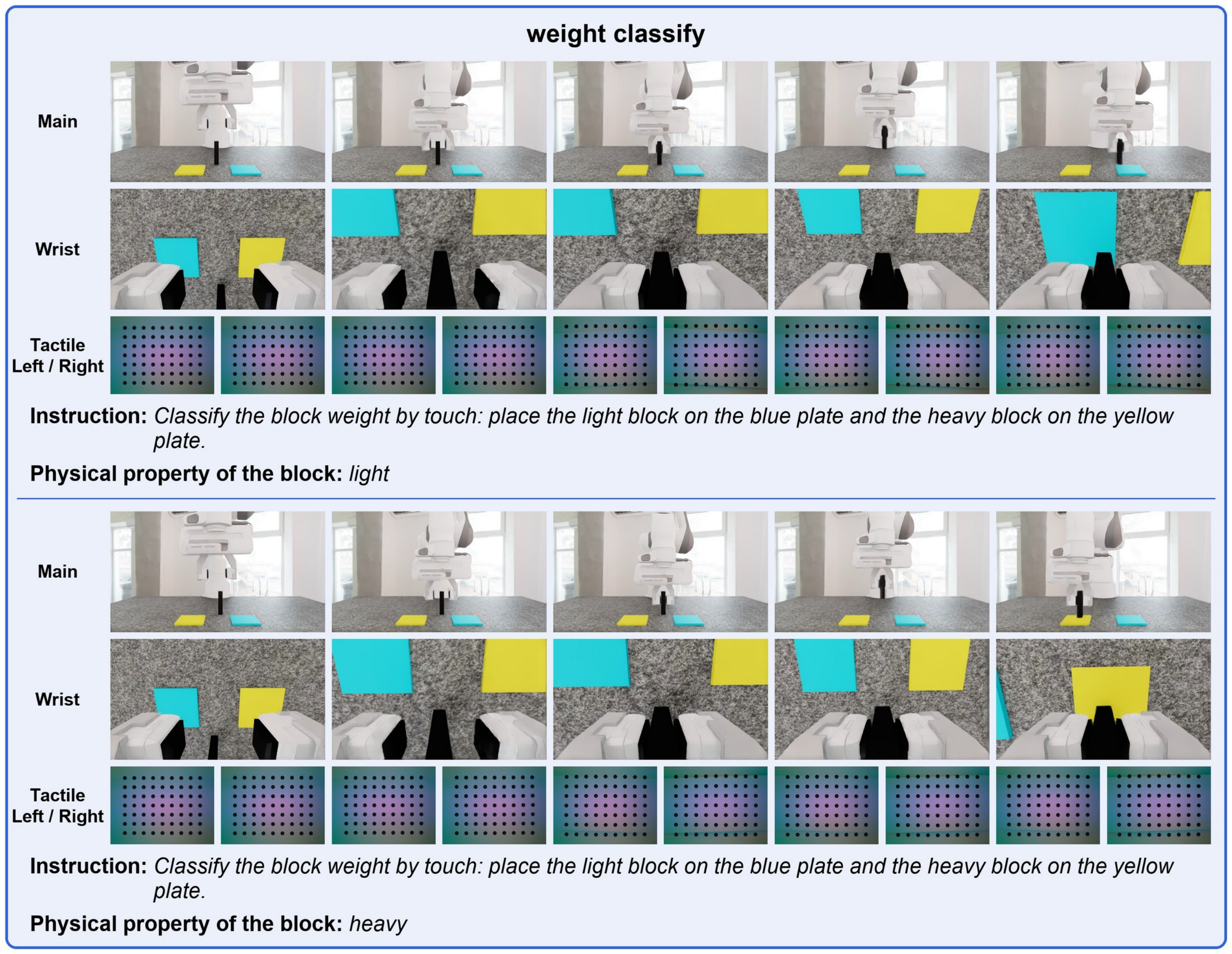}
  \caption{Visualization of \textit{weight classify} task.}
  \label{fig:weight_classify_visualization}
\end{figure}

\begin{figure}[H]
  \centering
  \includegraphics[width=0.8\linewidth]{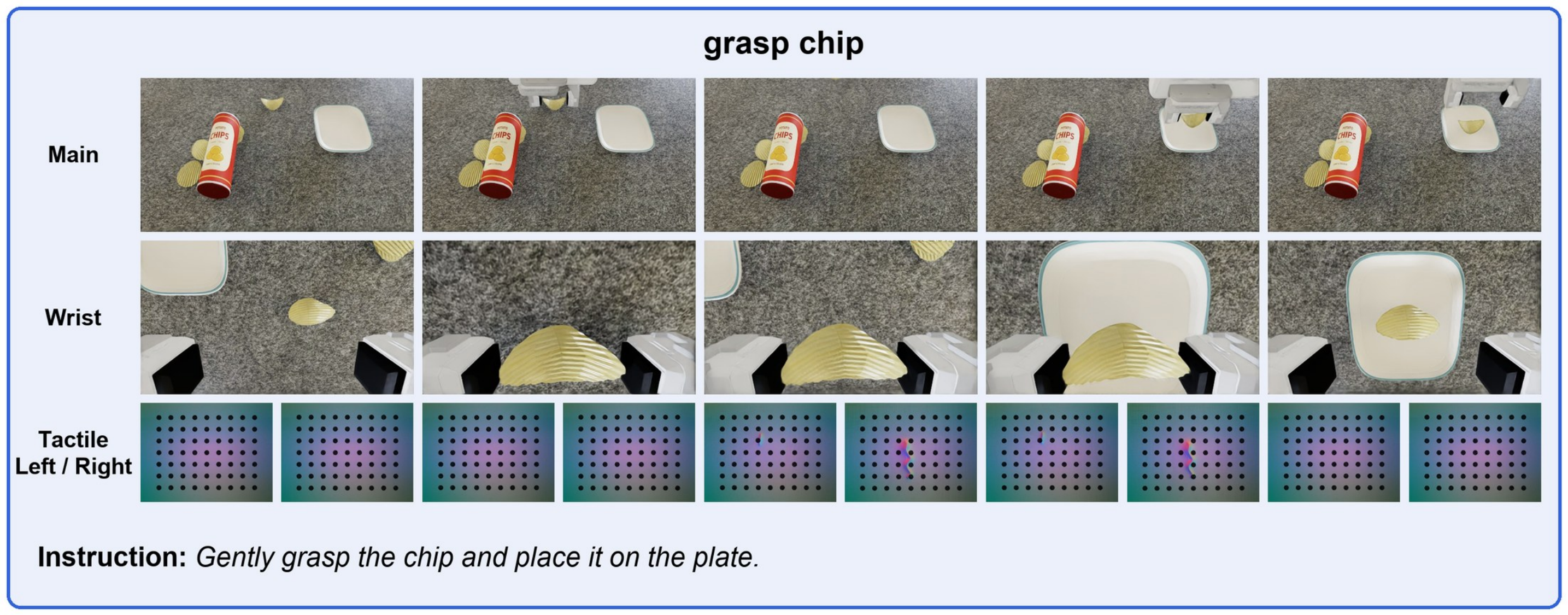}
  \caption{Visualization of \textit{grasp chip} task.}
  \label{fig:grasp_chip_visualization}
\end{figure}

\begin{figure}[H]
  \centering
  \includegraphics[width=0.8\linewidth]{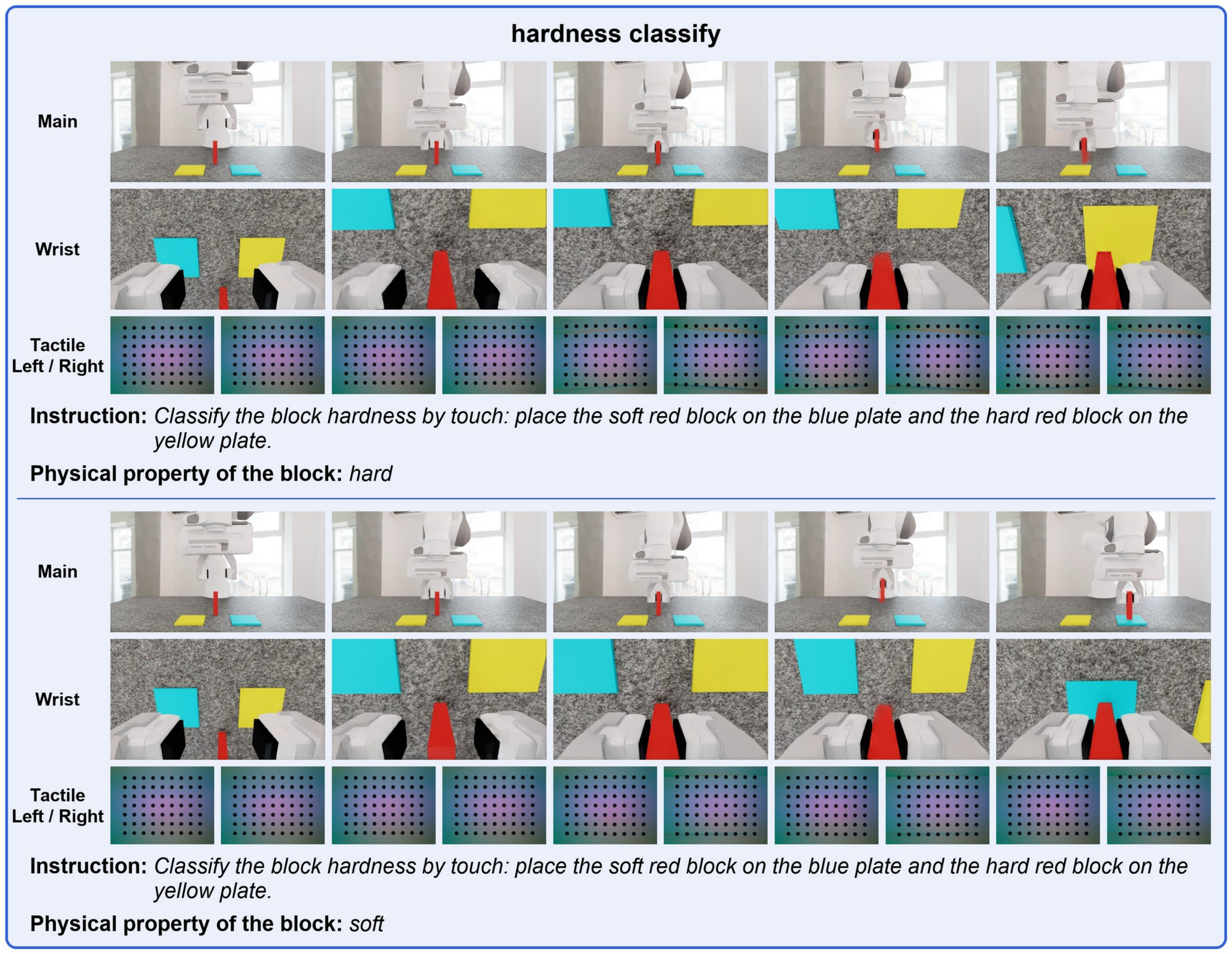}
  \caption{Visualization of \textit{hardness classify} task.}
  \label{fig:hardness_classify_visualization}
\end{figure}

\begin{figure}[H]
  \centering
  \includegraphics[width=0.8\linewidth]{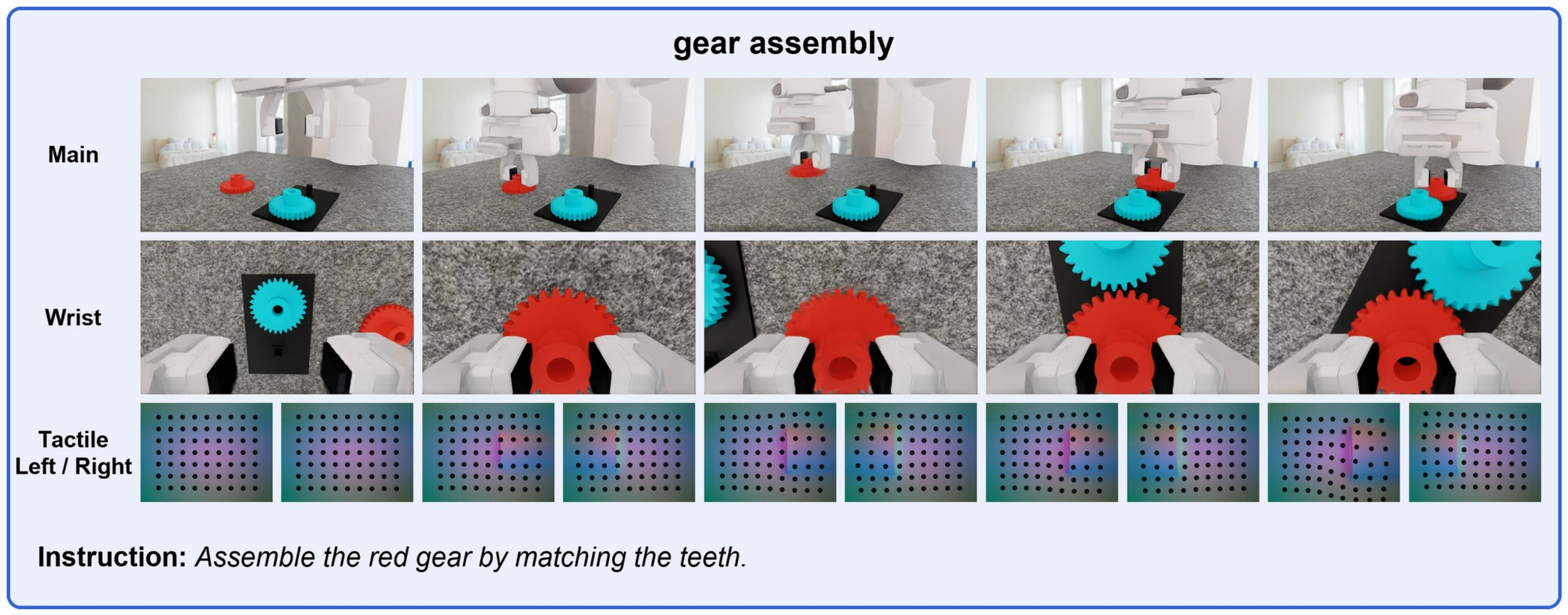}
  \caption{Visualization of \textit{gear assembly} task.}
  \label{fig:gear_assembly_visualization}
\end{figure}

\begin{figure}[H]
  \centering
  \includegraphics[width=0.8\linewidth]{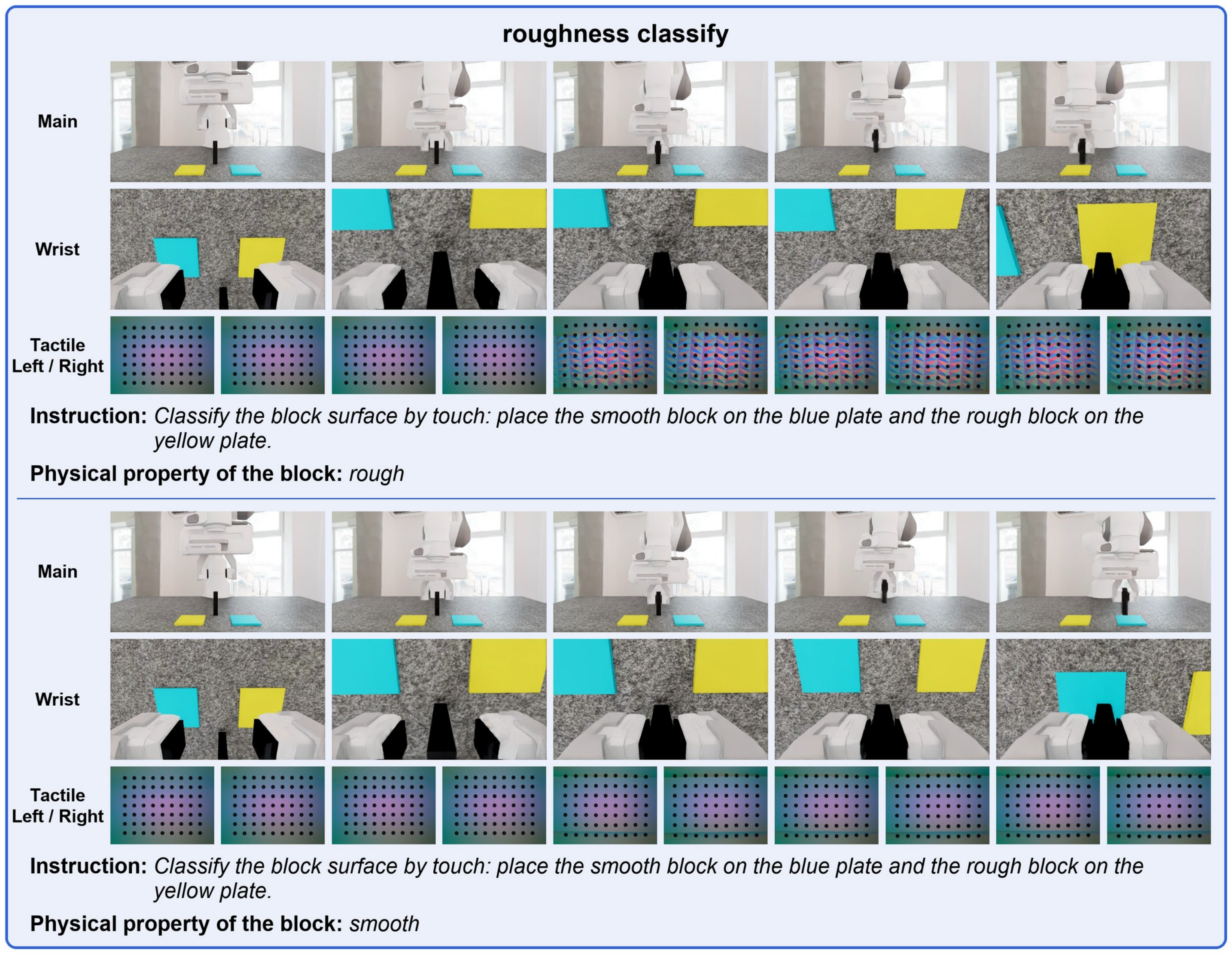}
  \caption{Visualization of \textit{roughness classify} task.}
  \label{fig:roughness_classify_visualization}
\end{figure}

\begin{figure}[H]
  \centering
  \includegraphics[width=0.8\linewidth]{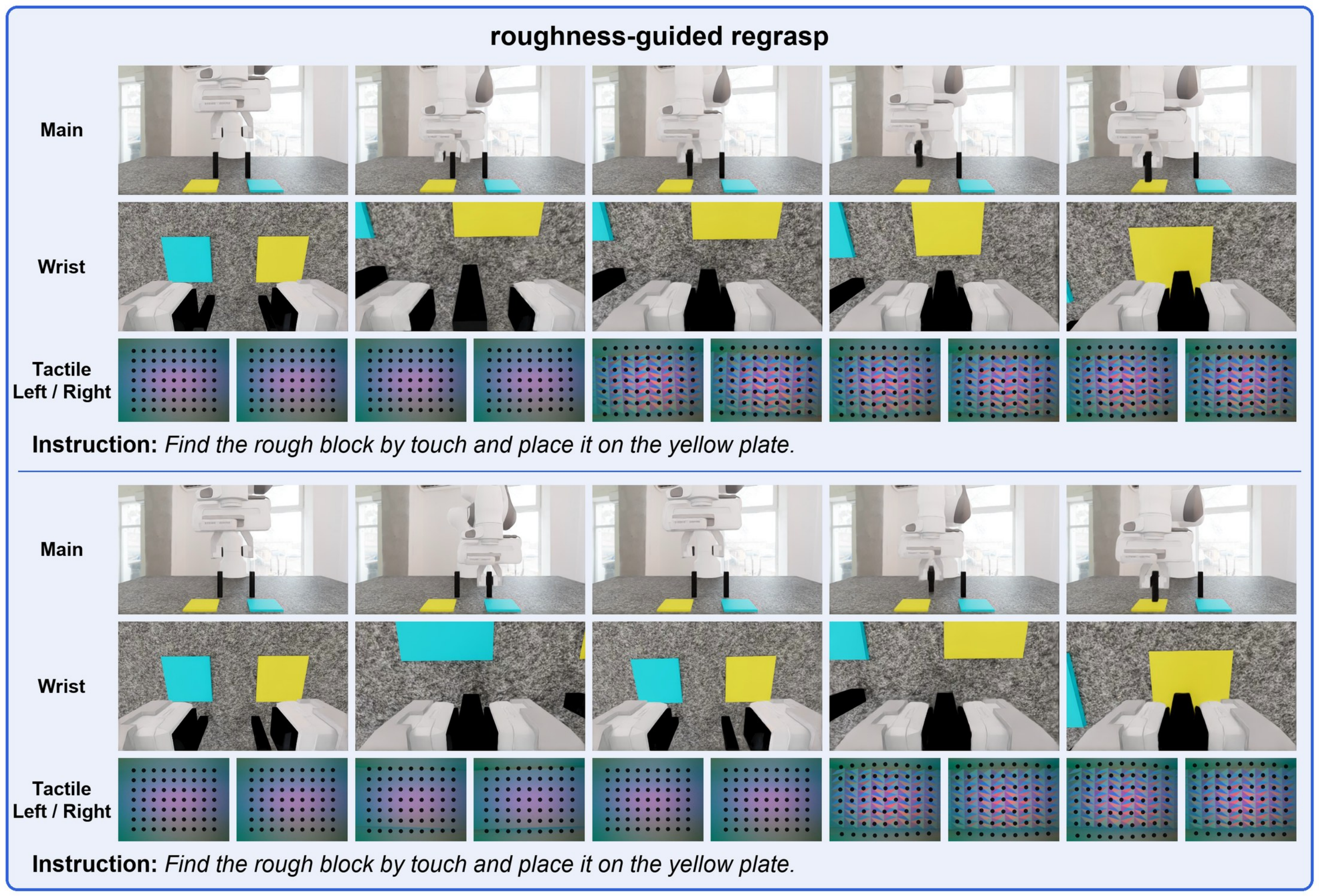}
  \caption{Visualization of \textit{roughness-guided regrasp} task.}
  \label{fig:roughness_regrasp_visualization}
\end{figure}

\begin{figure}[H]
  \centering
  \includegraphics[width=0.8\linewidth]{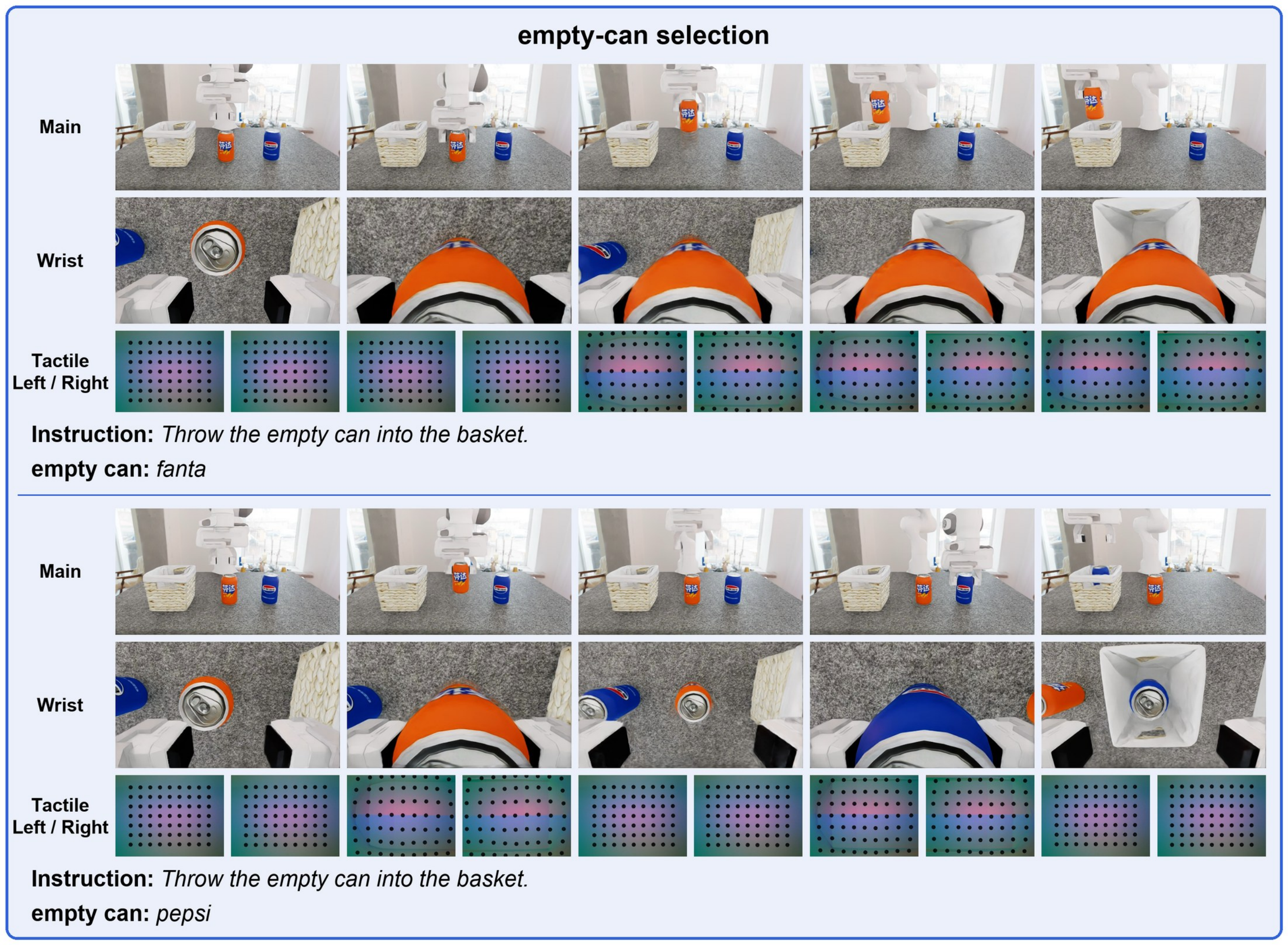}
  \caption{Visualization of \textit{empty-can selection} task.}
  \label{fig:empty_can_selection_visualization}
\end{figure}

\begin{figure}[H]
  \centering
  \includegraphics[width=0.8\linewidth]{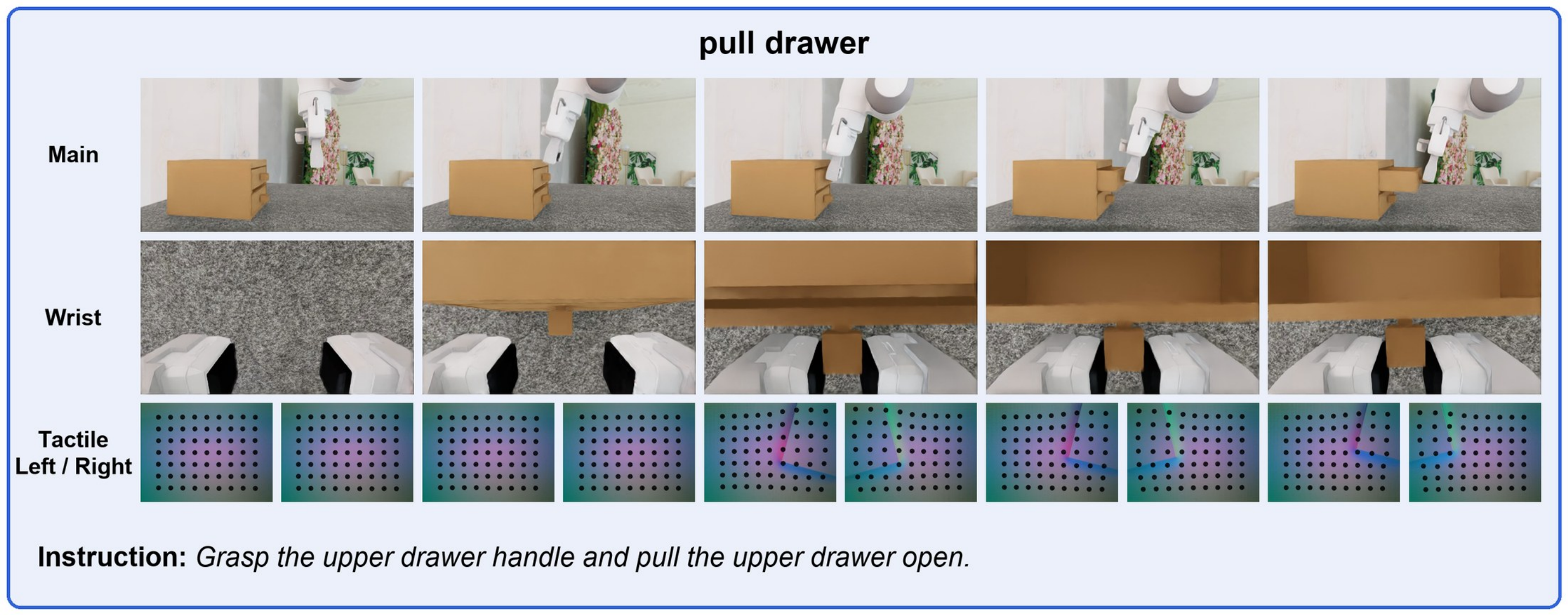}
  \caption{Visualization of \textit{pull drawer} task.}
  \label{fig:pull_drawer_visualization}
\end{figure}

\begin{figure}[H]
  \centering
  \includegraphics[width=0.8\linewidth]{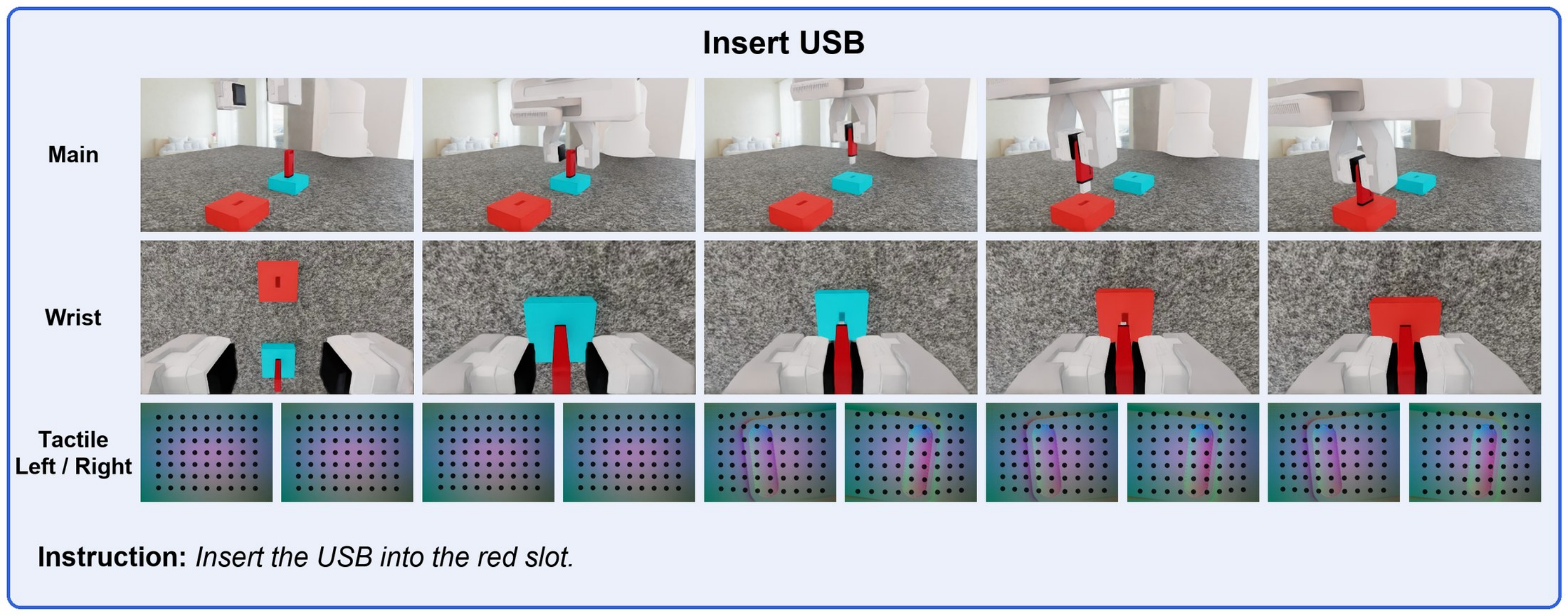}
  \caption{Visualization of \textit{insert USB} task.}
  \label{fig:insert_USB_visualization}
\end{figure}

\begin{figure}[H]
  \centering
  \includegraphics[width=0.8\linewidth]{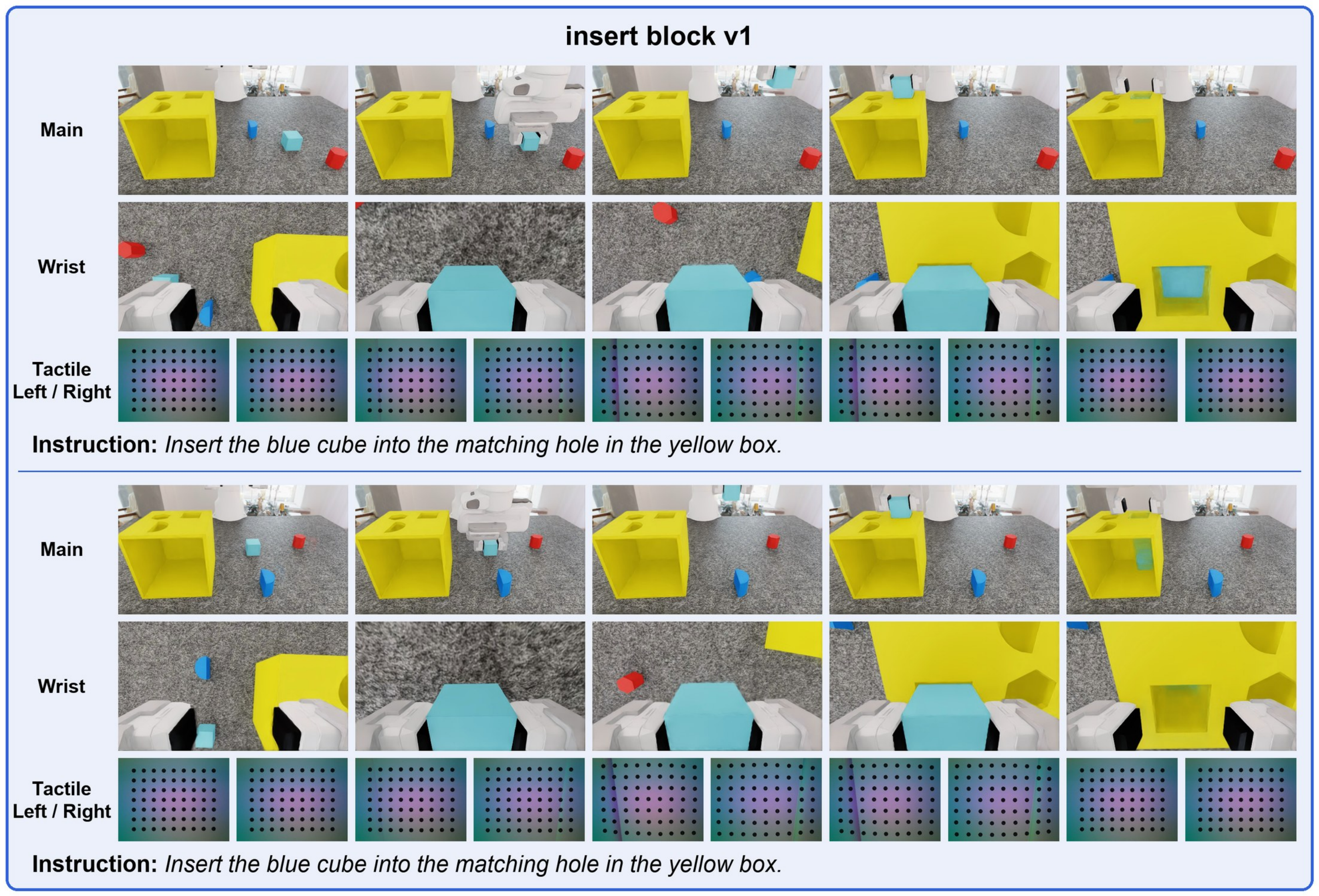}
  \caption{Visualization of \textit{insert block v1} task.}
  \label{fig:insert_block_v1_visualization}
\end{figure}

\begin{figure}[H]
  \centering
  \includegraphics[width=0.8\linewidth]{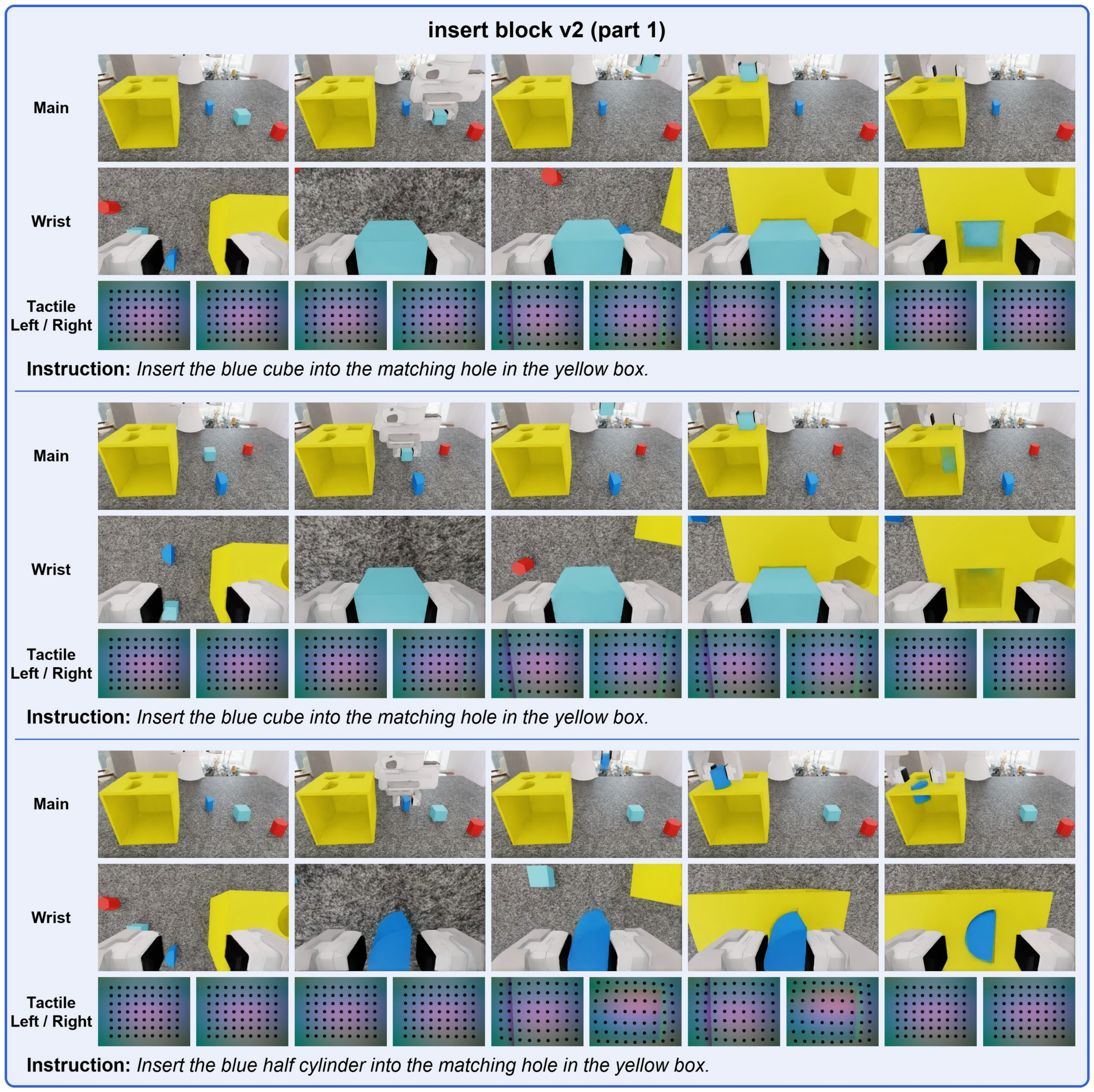}
  \caption{Visualization of \textit{insert block v2} task (part 1).}
  \label{fig:insert_block_v2_part1_visualization}
\end{figure}

\begin{figure}[H]
  \centering
  \includegraphics[width=0.8\linewidth]{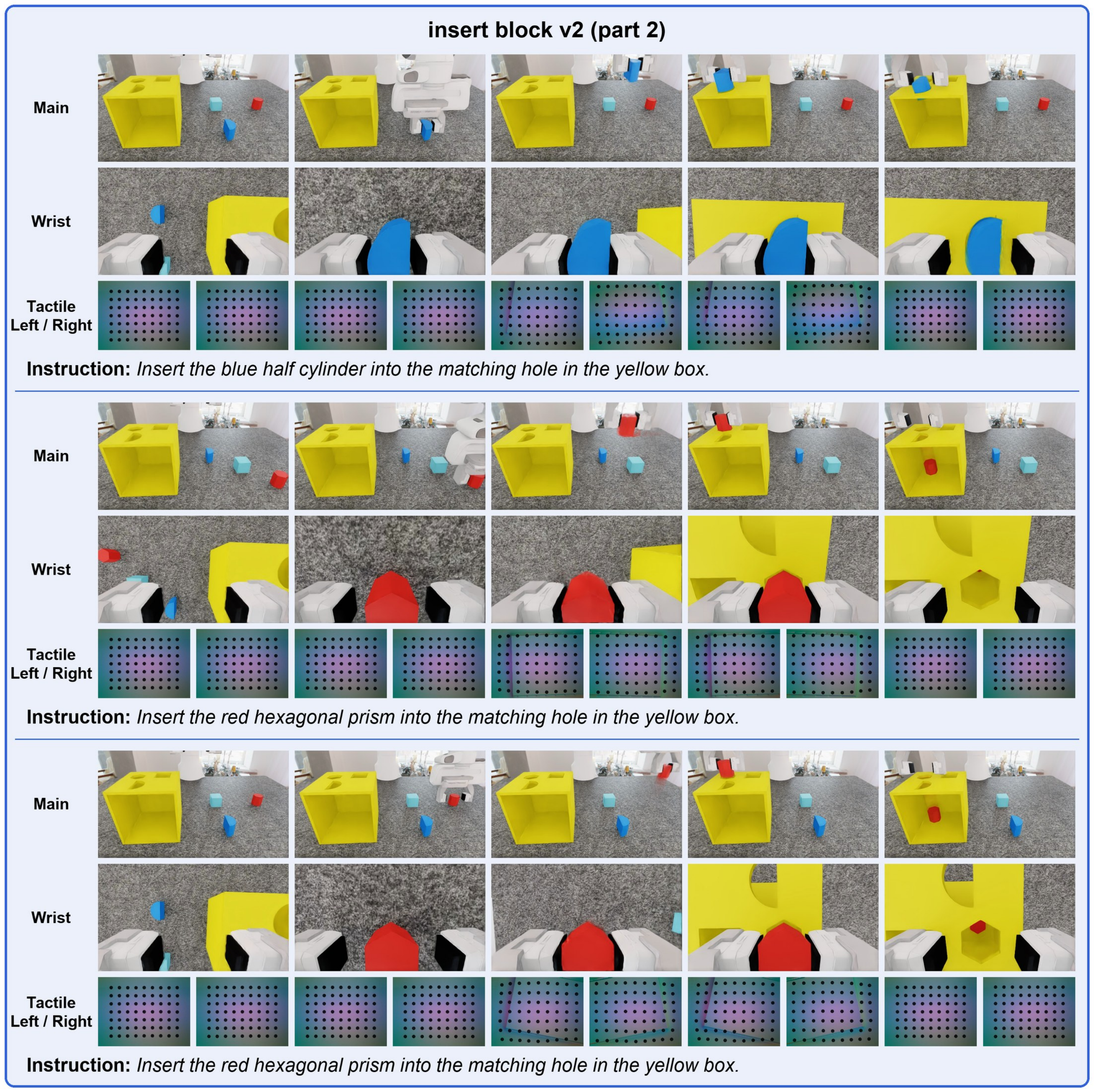}
  \caption{Visualization of \textit{insert block v2} task (part 2).}
  \label{fig:insert_block_v2_part2_visualization}
\end{figure}

Fig.~\ref{fig:weight_classify_visualization}--\ref{fig:insert_block_v2_part2_visualization} visualize all 11 simulated tasks across physical-property perception, fragility-aware manipulation, contact-rich manipulation, and precision manipulation.
The panels place main-camera, wrist-camera, and bilateral tactile observations in temporal order; multi-case figures additionally show distinct task conditions or object variants.
These examples make the interaction sequence and the contact information available to the policy easier to inspect.

\clearpage
\subsection{Real-World Tasks}

\begin{figure}[H]
  \centering
  \includegraphics[width=0.8\linewidth]{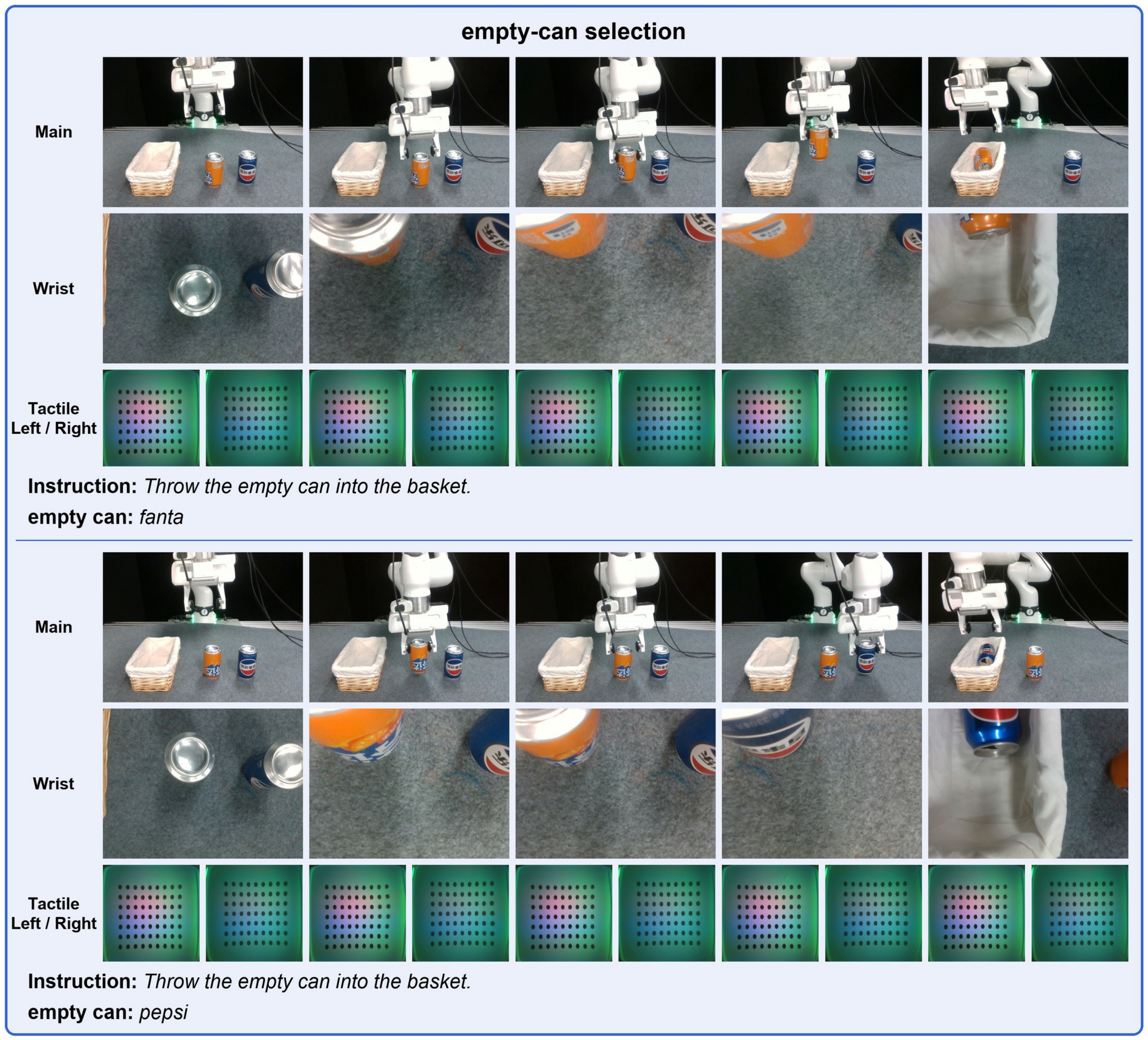}
  \caption{Visualization of \textit{empty-can selection} task.}
  \label{fig:real_empty_can_selection_visualization}
\end{figure}

\begin{figure}[H]
  \centering
  \includegraphics[width=0.8\linewidth]{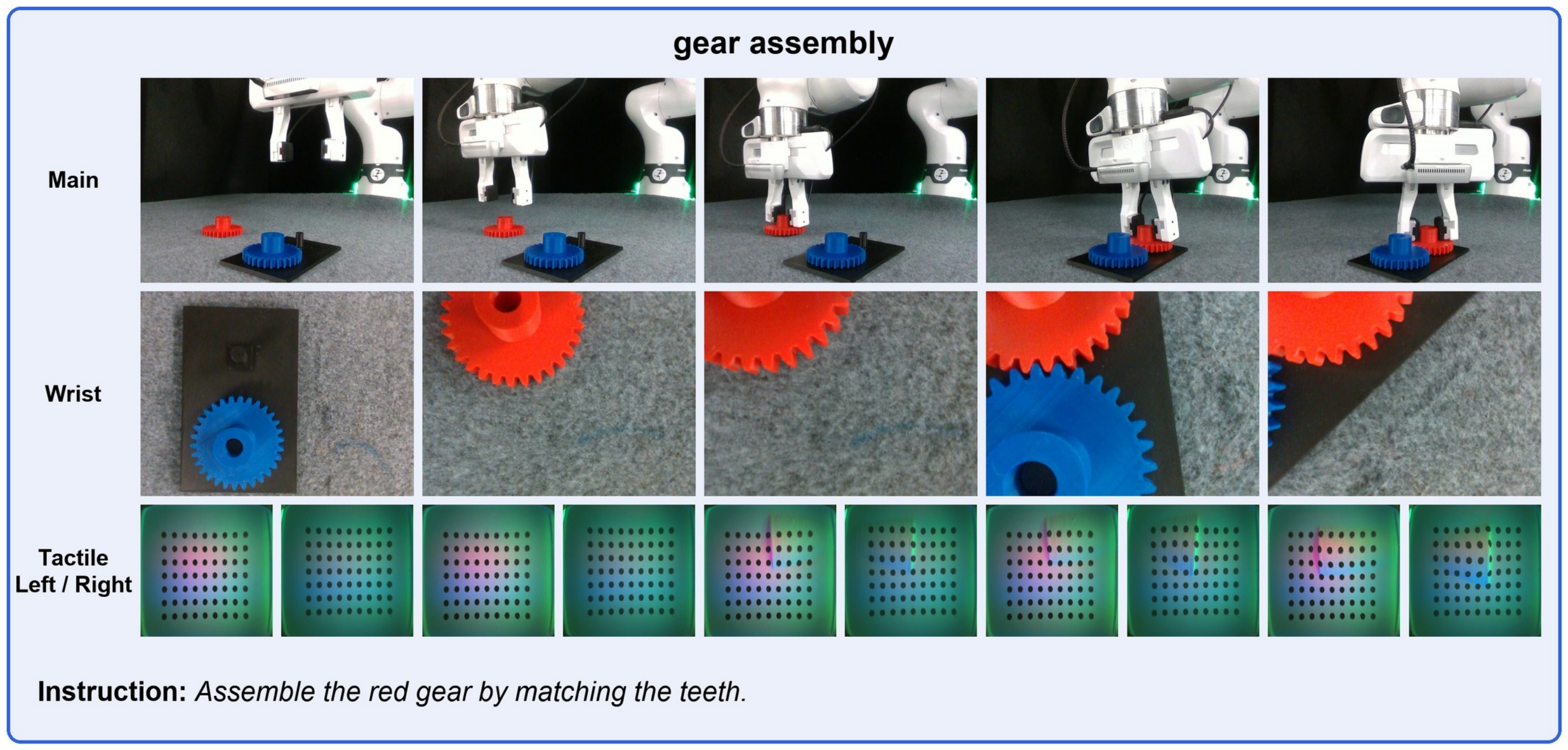}
  \caption{Visualization of \textit{gear assembly} task.}
  \label{fig:real_gear_assembly_visualization}
\end{figure}

\begin{figure}[H]
  \centering
  \includegraphics[width=0.8\linewidth]{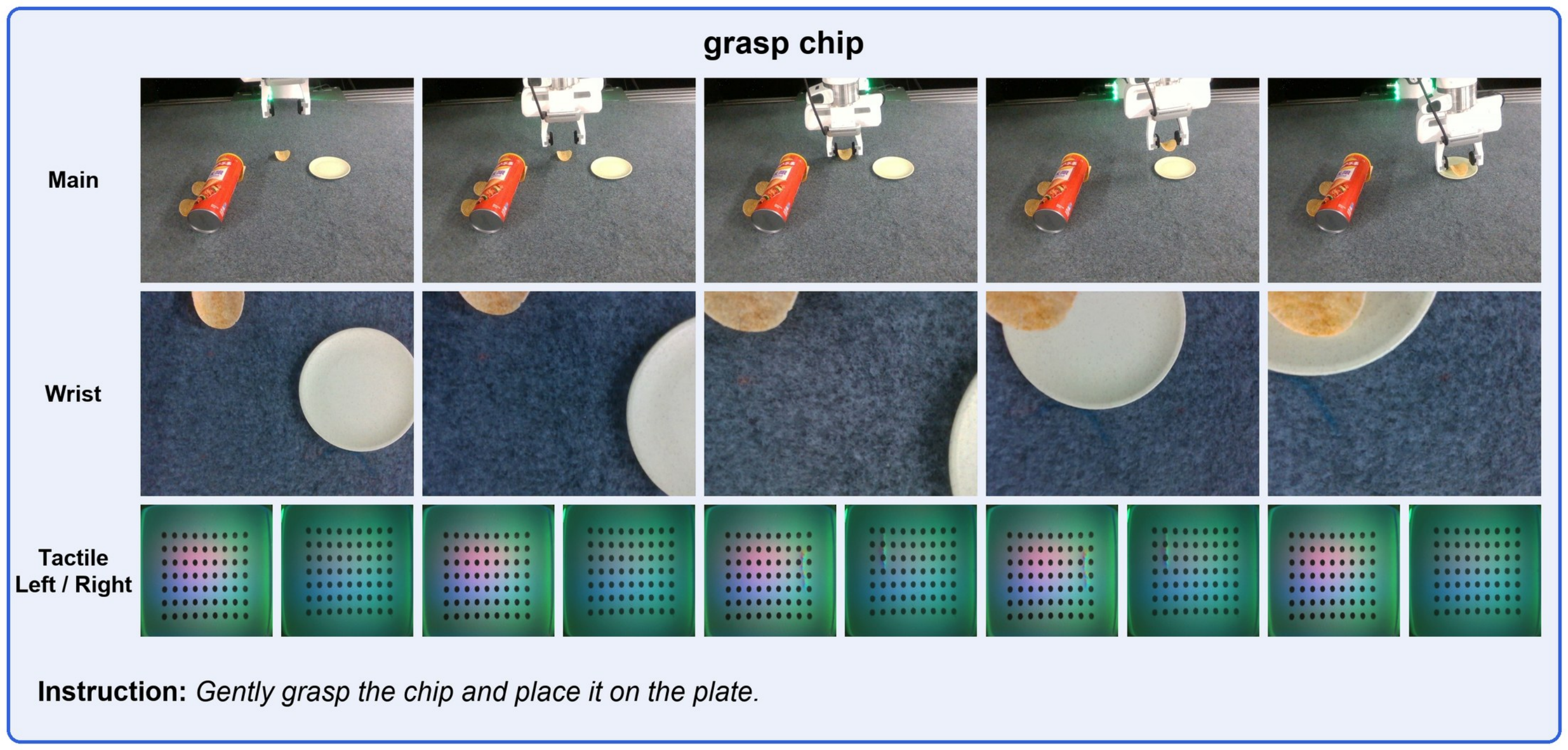}
  \caption{Visualization of \textit{grasp chip} task.}
  \label{fig:real_grasp_chip_visualization}
\end{figure}

\begin{figure}[H]
  \centering
  \includegraphics[width=0.8\linewidth]{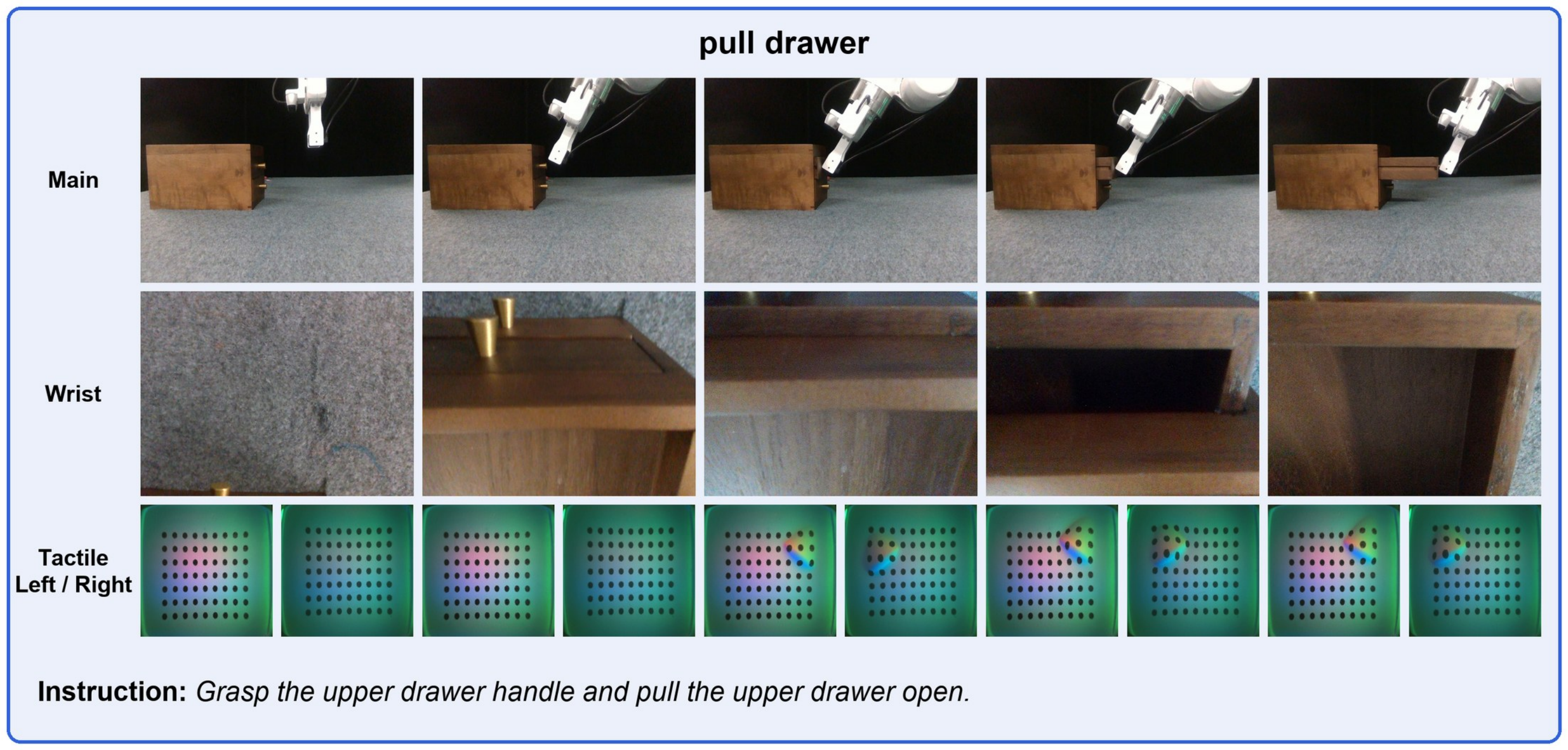}
  \caption{Visualization of \textit{pull drawer} task.}
  \label{fig:real_pull_drawer_visualization}
\end{figure}

\begin{figure}[H]
  \centering
  \includegraphics[width=0.8\linewidth]{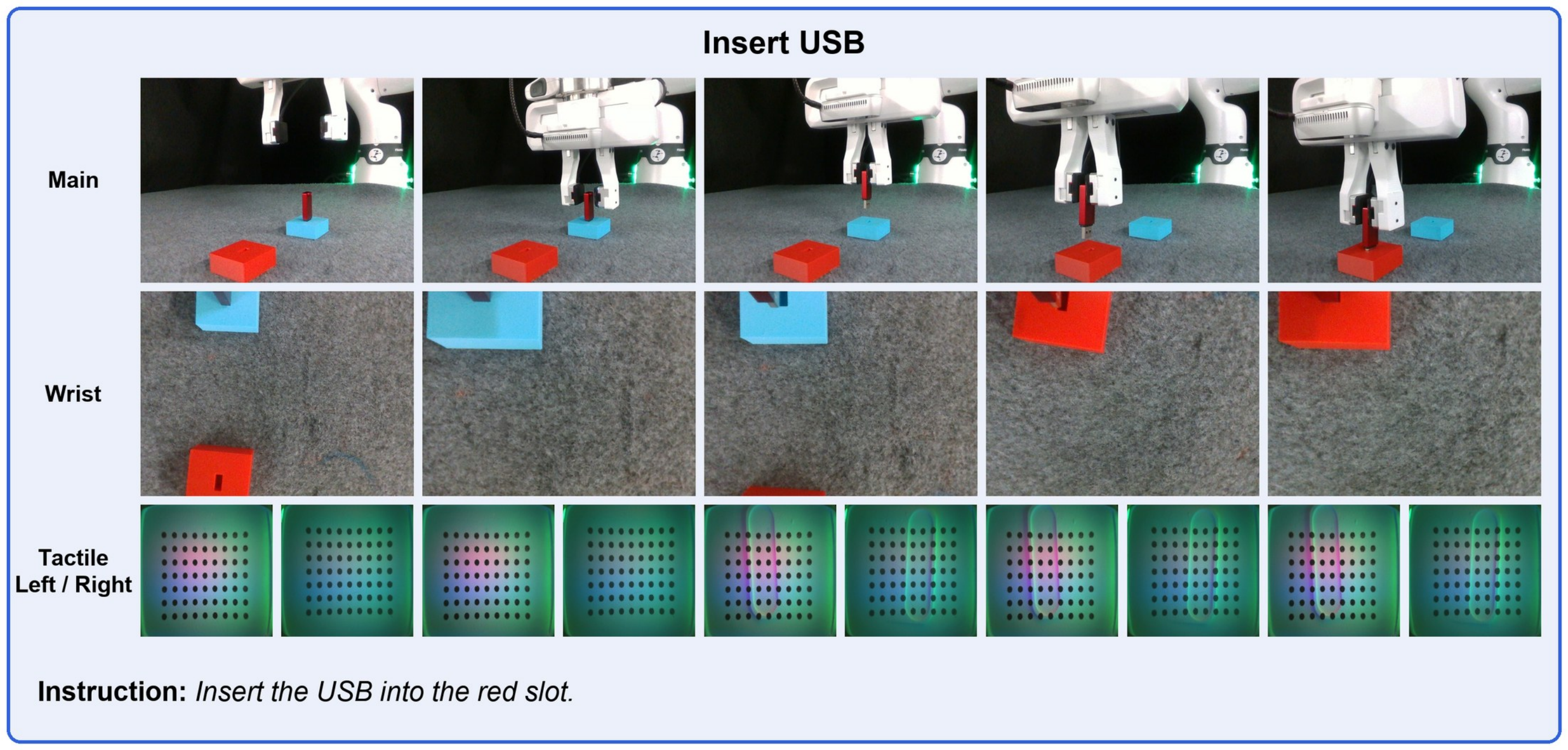}
  \caption{Visualization of \textit{insert USB} task.}
  \label{fig:real_insert_USB_visualization}
\end{figure}

\begin{figure}[H]
  \centering
  \includegraphics[width=0.8\linewidth]{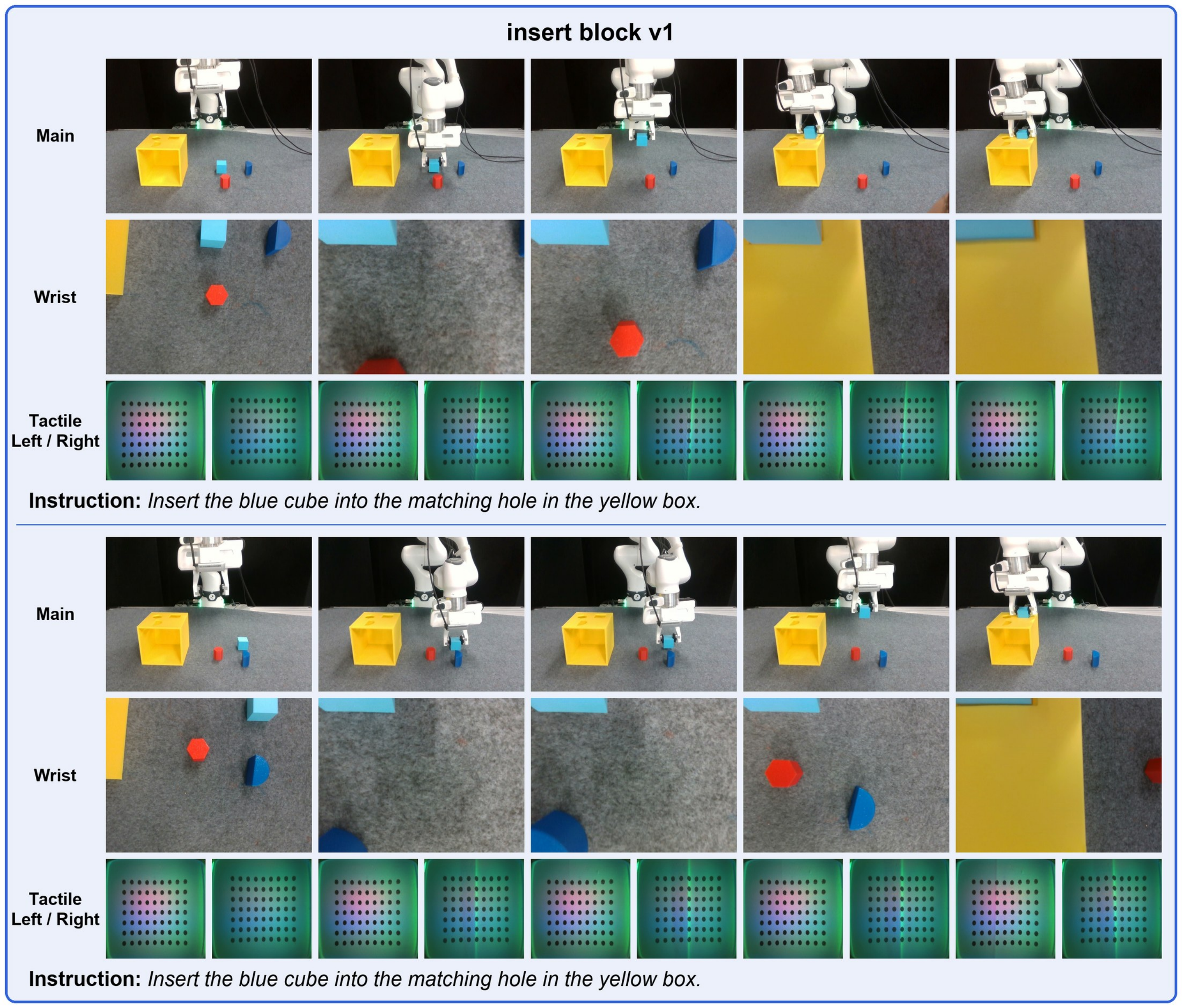}
  \caption{Visualization of \textit{insert block v1} task.}
  \label{fig:real_insert_block_v1_visualization}
\end{figure}

\begin{figure}[H]
  \centering
  \includegraphics[width=0.8\linewidth]{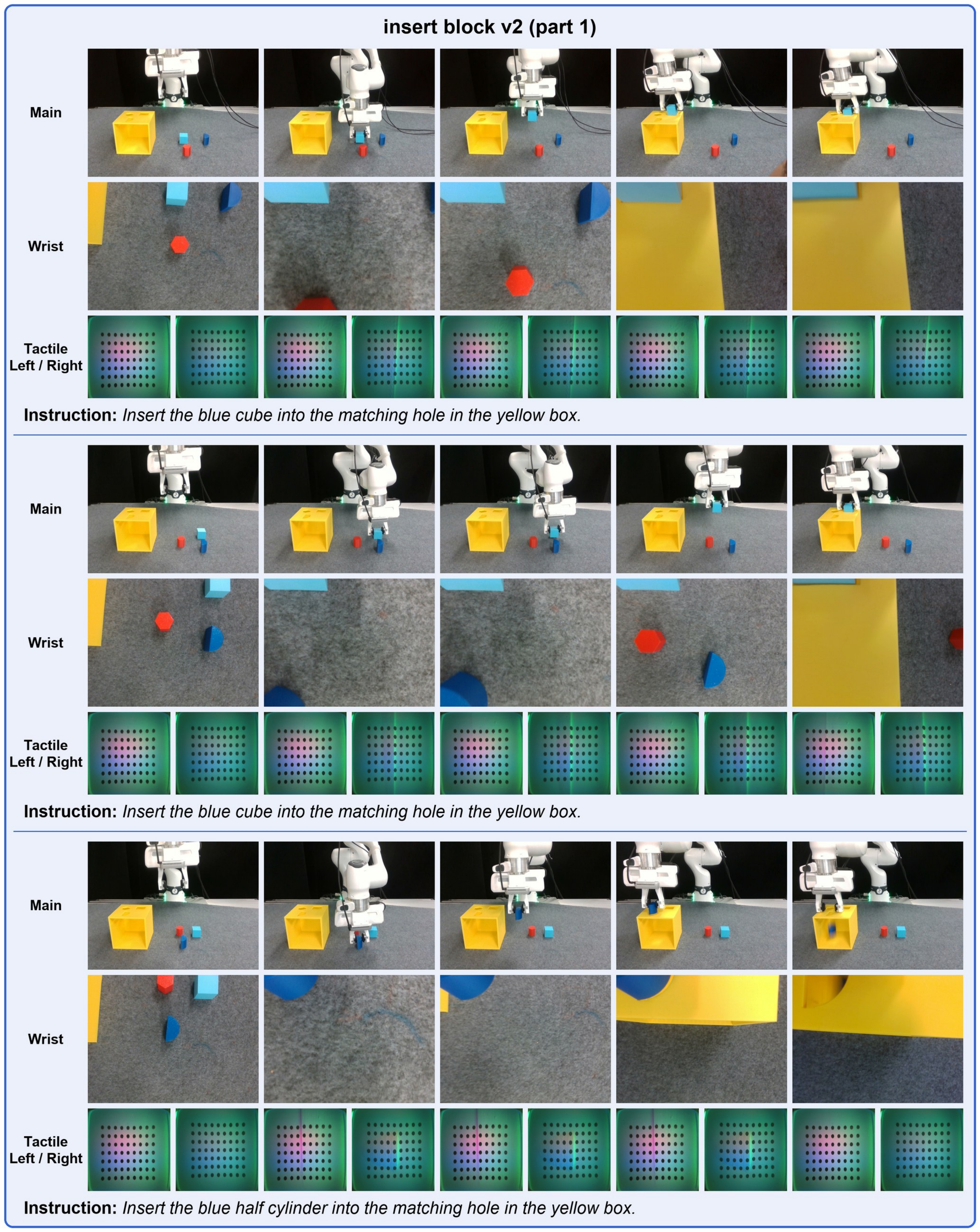}
  \caption{Visualization of \textit{insert block v2} task (part 1).}
  \label{fig:real_insert_block_v2_part1_visualization}
\end{figure}

\begin{figure}[H]
  \centering
  \includegraphics[width=0.8\linewidth]{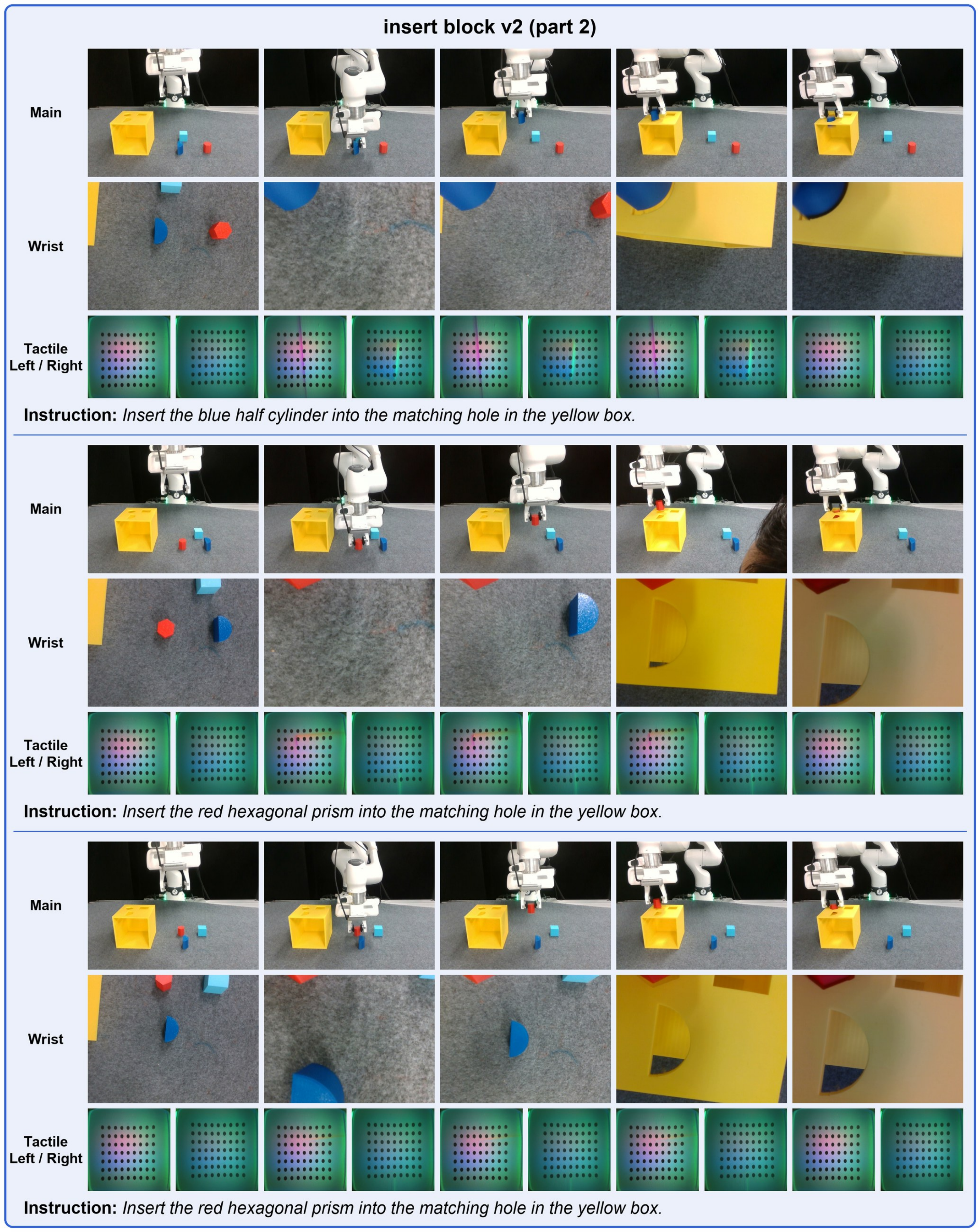}
  \caption{Visualization of \textit{insert block v2} task (part 2).}
  \label{fig:real_insert_block_v2_part2_visualization}
\end{figure}

Fig.~\ref{fig:real_empty_can_selection_visualization}--\ref{fig:real_insert_block_v2_part2_visualization} show representative real-world instances of the paired benchmark tasks.
As in the simulated figures, the camera and fingertip tactile views are arranged over successive interaction stages.
The examples provide qualitative context for object selection, gentle grasping, sustained contact, and precise insertion with physical objects and sensors.

\end{document}